\documentclass[10pt,twocolumn,letterpaper]{article}

\usepackage{cvpr}              

\usepackage[dvipsnames]{xcolor}

\usepackage{url}
\usepackage{cite}
\usepackage{amsmath,amssymb,amsfonts}
\usepackage{algorithmic}
\usepackage{graphicx}
\usepackage{textcomp}
\usepackage{paralist}
\usepackage{listings}
\usepackage{booktabs}
\usepackage{hhline}
\usepackage{framed,multirow}
\usepackage{lipsum}  
\usepackage{soul}
\usepackage{cancel}
\usepackage{booktabs}
\usepackage{latexsym}
\usepackage{array}
\usepackage{stackengine}
\usepackage{relsize}
\usepackage{multirow}
\usepackage[ruled,vlined]{algorithm2e}
\usepackage[dvipsnames]{xcolor}

\definecolor{cvprblue}{rgb}{0.21,0.49,0.74}
\usepackage[pagebackref,break links,colorlinks,citecolor=cvprblue]{hyperref}

\newcommand{\x}{\mathbf{x}}

\newcommand{\z}{\mathbf{z}}
\newcommand{\h}{\mathbf{h}}
\newcommand{\largeh}{\mathbf{H}}

\newcommand{\ours}{\textsc{Panther}}

\newcommand{\slide}{\mathbf{z}_{\text{WSI}}^j}

\newcommand{\supp}{\textbf{Supplementary Information}}

\makeatletter
\renewcommand*{\@fnsymbol}[1]{%
  \ensuremath{%
    \ifcase#1\or *
    \or \dagger
    \or \ddagger 
    \or \mathsection
    \or \mathparagraph
    \or \|\or **\or \dagger\dagger
    \or \ddagger\ddagger \else\@ctrerr\fi}%
}
\makeatother

\def\paperID{10434} 
\def\confName{CVPR}
\def\confYear{2024}

\title{Morphological Prototyping for Unsupervised Slide Representation Learning\\ in Computational Pathology}

\author{Andrew H. Song$^{1,2,}$\thanks{Equal contribution} , Richard J. Chen$^{1,2,\ast}$, Tong Ding$^{1,2}$, Drew F.K. Williamson$^{1,2,}$\thanks{Presently at Emory University School of Medicine} , \\
Guillaume Jaume$^{1,2}$, Faisal Mahmood$^{1,2}$\\
${^1}$Mass General Brigham and ${^2}$Harvard University\\
{\tt\small asong@bwh.harvard.edu, richardchen@g.harvard.edu, faisalmahmood@bwh.harvard.edu}
}
\begin{document}
\maketitle

\begin{abstract}
Representation learning of pathology whole-slide images (WSIs) has been has primarily relied on weak supervision with Multiple Instance Learning (MIL). However, the slide representations resulting from this approach are highly tailored to specific clinical tasks, which limits their expressivity and generalization, particularly in scenarios with limited data. Instead, we hypothesize that morphological redundancy in tissue can be leveraged to build a task-agnostic slide representation in an unsupervised fashion. To this end, we introduce $\ours$, a prototype-based approach rooted in the Gaussian mixture model that summarizes the set of WSI patches into a much smaller set of morphological prototypes. Specifically, each patch is assumed to have been generated from a mixture distribution, where each mixture component represents a morphological exemplar. Utilizing the estimated mixture parameters, we then construct a compact slide representation that can be readily used for a wide range of downstream tasks.
By performing an extensive evaluation of $\ours$ on subtyping and survival tasks using 13 datasets, we show that 1) $\ours$ outperforms or is on par with supervised MIL baselines and 2) the analysis of morphological prototypes brings new qualitative and quantitative insights into model interpretability. The code is available at \url{https://github.com/mahmoodlab/Panther}.
\end{abstract}    
\section{Introduction}
\label{sec:intro}

Representation learning of whole-slide images (WSIs) is a fundamental task in Computational Pathology (CPath)~\cite{song2023artificial}. Given a WSI, the goal is to learn a slide-level representation that can be used for various downstream tasks, such as diagnosis, prognostication, and therapeutic response prediction. The standard approach is weakly supervised learning based on Multiple Instance Learning (MIL)~\cite{dundar2010multiple, ilse2018attention,campanella2019clinical}, in which the gigapixel WSI is tokenized into a large set of patch embeddings ($N>10,000$) with a pretrained vision encoder, followed by aggregation of the embeddings~\cite{ilse2018attention}. Current advances in CPath have examined: (1) learning better patch embeddings with domain-specific vision encoders based on self-supervised learning (SSL)~\cite{ciga2022self,wang2022transformer, Filiot2023Scaling, azizi2023robust, kang2023benchmarking,jiang2023hierarchical, huang2023visual, chen2024towards, lu2024visual} and (2) developing new aggregation strategies for pooling patch embeddings into a slide representation~\cite{li2021dual, lu2021data, shao2021transmil, lin2023interventional}. As many histology datasets have limited samples for supervised MIL, an emerging goal in CPath is to (3) shift slide representation learning from \emph{weakly-supervised} to \emph{unsupervised}~\cite{chen2022scaling, lazard2023giga,VU2023handcrafted,quiros2023mapping,jaume2024transcriptomics}, which may help mitigating data and label scarcity and improving generalization.

\begin{figure}[t]
   \centering
   \includegraphics[width=1\linewidth]{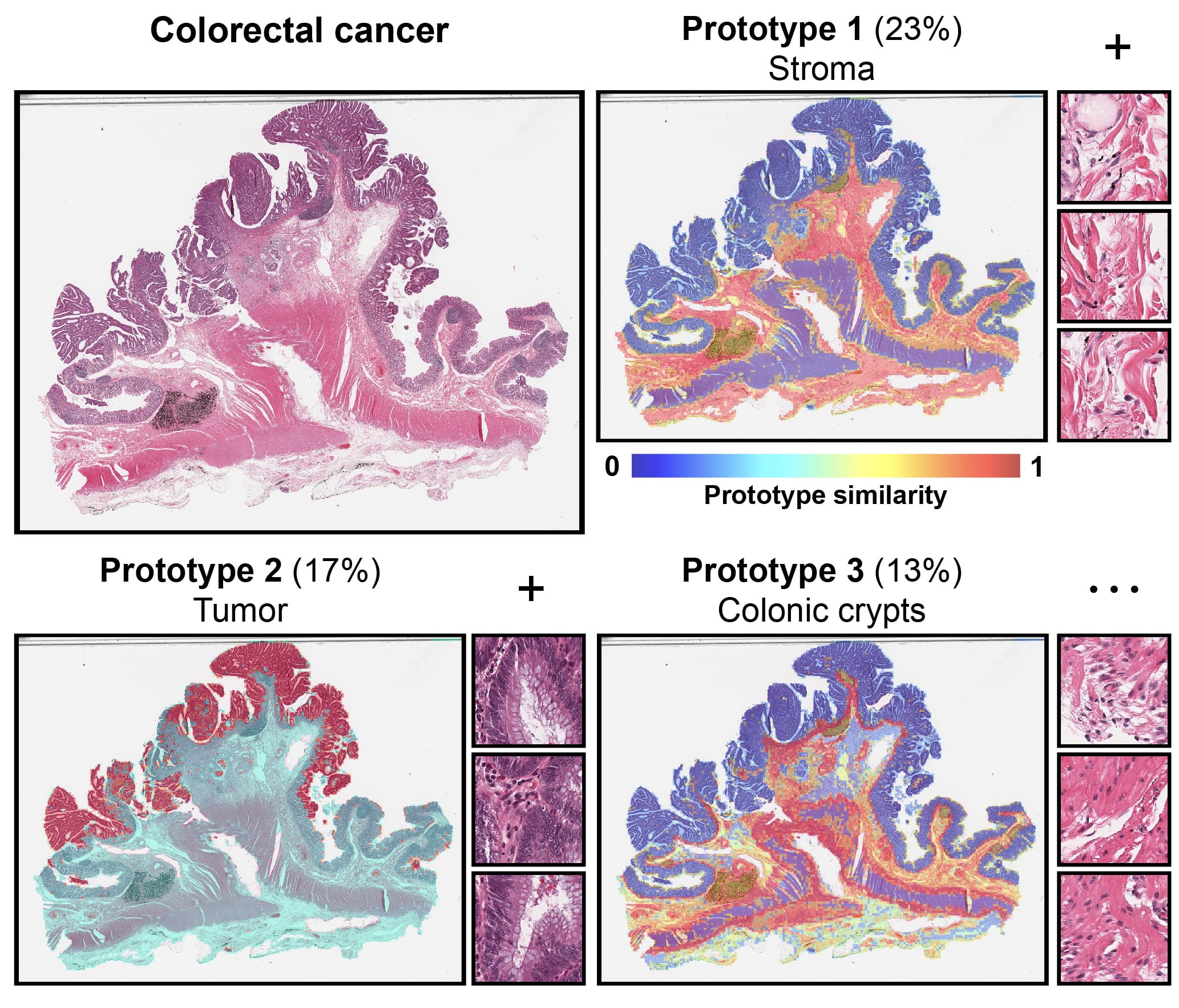}
   \caption{\textbf{Slide decomposition into morphological prototypes} Due to morphological redundancy across and within the tissue, a slide can be decomposed into prototypes. We introduce $\ours$, a method that can identify and extract morphological prototypes to form a compact and unsupervised slide representation.
   }
   \label{fig:top}
\end{figure}

We postulate that such models are particularly suited for fine-grained classification tasks, such as survival outcome prediction that require holistic modeling of morphologies found in the tissue microenvironment~\cite{yao2020whole,zhu2017wsisa}.
In contrast to ``needle-in-a-haystack" tasks (\emph{e.g.,} micro-metastasis detection) that require localizing tumor patches, ``panoramic" tasks require integrating spatial heterogeneity (diversity of distinct tumor populations)~\cite{heindl2015mapping,lu2020capturing}, interactions and context (immune infiltrate near invasive tumor margin)~\cite{beck2011systematic, saltz2018spatial}, and size (number and size of masses)~\cite{brierley2017tnm}. Attention-based architectures~\cite{ilse2018attention, lu2021data, javed2022additive} demonstrate clinical-grade performance in the former task (selectively focusing on diagnostic patches of a single visual concept)~\cite{bejnordi2017diagnostic}; however, they have limited expressivity in the panoramic tasks that benefit from understanding proportions and mixtures of visual concepts~\cite{chen2022towards, bueno2021representation,chan2019histosegnet}. 

Based on these insights, we propose an \textit{unsupervised} slide representation framework that can accurately capture the proportions and mixtures of morphological visual concepts. Specifically, we build on the observation that WSI patches show \textit{morphological redundancy} and thus a handful of morphological patterns are repeated (\textbf{Fig.}~\ref{fig:top}). Formally, we hypothesize that a concise set of key descriptors (prototypes), coupled with distribution characterizing the extent and variation of each descriptor, would comprehensively summarize the WSI.  
As this summary only relies on the statistical characteristics of each WSI, this yields a generic and unbiased slide embedding applicable across multiple downstream tasks. To faithfully summarize the WSI, the goal becomes to construct (1) a mapping between each patch and the prototypes and (2) a slide embedding with the learned mapping that includes representation of each prototype and its extent, \emph{i.e.,} its cardinality.

To this end, we introduce $\ours$, a \textbf{P}rototype \textbf{A}ggregatio\textbf{N}-based framework for compac\textbf{T} \textbf{HE}terogenous slide set \textbf{R}epresentation ($\ours$). Inspired by previous work in prototype-based set representation learning~\cite{mialon2021a, guo2022learning, kim2022differentiable, ye2024ptarl}, $\ours$ builds an \textit{unsupervised} slide embedding by assuming that each patch embedding is generated from the Gaussian mixture model (GMM), with each morphological prototype representing a mixture component. By employing GMM, the two aforementioned goals are seamlessly satisfied through the parameter estimation procedure with Expectation-Maximization (EM)~\cite{dempster1977maximum, kim2022differentiable}: (1) the estimated posterior probability of mixture assignment for each patch defines the mapping between a patch embedding and a prototype, and (2) mixture probability represents cardinality, and mixture mean \& covariance represent the representation of corresponding morphological pattern. The slide representation is formed as a concatenation of the estimated GMM parameters across all prototypes. Since the representation can be decomposed along each prototype, this allows for per-prototype nonlinear modeling and interpretation of the slide centered around each histology visual concept. 

To summarize, our contributions are (1) the first prototypical framework for learning compact and unsupervised \textit{slide representations} in WSI based on GMM; (2) a comprehensive evaluation of four diagnostic and nine prognostic tasks, demonstrating the outperformance against nearly all unsupervised and supervised baselines; (3) post-hoc interpretability with the quantification and visualization of morphological prototype spread within the tissue.

\section{Related work}
\label{sec:previous}
\subsection{Multiple instance learning in CPath}
While initial histology-based outcome prediction was centered on pathologist-annotated region-of-interests~\cite{mobadersany2018predicting, bychkov2018deep, kather2019predicting}, later works have utilized WSIs for clinical prediction tasks with MIL~\cite{ilse2018attention,skrede2020deep, campanella2019clinical, coudray2018classification, chen2022scaling}. There is a sustained effort for new MIL schemes, with a focus on new patch aggregation strategies to learn more representative and task-specific slide embedding, towards better predictive accuracy~\cite{shao2021transmil, li2021dual,xiang2023exploring,tang2023multiple,zhang2022dtfd,lu2023visual} or interpretability~\cite{javed2022additive, thandiackal2022differentiable}. MIL methods based on multiscale representation slides have recently shown promise for ``panoramic'' tasks~\cite{hashimoto2020multi, hou2022h, bontempo2023mil}.
$\ours$ is similar to MIL in that the slide-level embedding is constructed from patches for outcome prediction. However, $\ours$ constructs an \textit{unsupervised} set embedding and is agnostic to downstream tasks, in contrast to supervised MIL frameworks.

\subsection{Quantification of distance between sets}
Measuring the distance between two distinct sets (Wasserstein distance), \emph{e.g.,} between supports of empirical probability distributions, has received increasing attention from a diverse range of disciplines such as signal processing~\cite{Kolouri2017Optimal}, vision-language tasks~\cite{pramanick2023volta}, and computational biology~\cite{basu2014detecting, schiebinger2019optimal, bunne2023learning}. Given a similarity (or cost) metric between set elements such as $\mathcal{L}_2$ distance, the Wasserstein distance is defined as transporting (or matching) elements of one set to the other, incurring a minimum cost. Computing such distance is commonly called the optimal transport (OT)~\cite{cuturi2013sinkhorn}.


It is natural to extend this idea to CPath, where the biological entities of interest (e.g., WSI and genomic data) are typically modeled as a set of biological concepts. Different from MIL setting where a set of WSI patches is matched to a patient-level clinical label, this concerns matching between two sets: 1) sets of WSI patches for slide retrieval~\cite{cui2023retrievalaugmented} or domain adaptation~\cite{falahkheirkhah2023domain}, 2) different cancer datasets to quantify morphological distance between different cancer types~\cite{yeaton2022hierarchical}, and 3) a set of WSI patches and a set of genomic tokens to learn optimal fusion for improved prognosis~\cite{xu2023multimodal}. Similarly, $\ours$ can be seen as the matching problem between a set of WSI patches and prototypes. 

\subsection{Prototype-based set representation}
Prototypes are representative examples of data points that share the same class, usually formulated as centroids from clustering that describe unique human-interpretable concepts and other semantic information~\cite{chen2004image, snell2017prototypical, chen2019looks}. Recently, it has been applied to compactly represent large set data in bioinformatics and NLP~\cite{lee2019set, mialon2021a, kim2022differentiable, guo2022learning}. The desiderata for prototype-based set representation is to model: (1) cardinality, i.e., how many elements in the set are associated with a prototype, and (2) description, i.e., prototype identity.

Posed also in many related forms such as signatures~\cite{lazebnik2005sparse, zhang2007local} and bag of visual words (BoVW)~\cite{sivic2003video,cruz2011visual,caicedo2009histopathology}, learning prototypical representations is a natural problem in CPath as repeating histology patterns often reflect the same  morphology~\cite{vu2015histopathological,xu2012multiple,hou2016patch,yang2023tpmil,pan2023human,yu2023prototypical,kalra2020yottixel}. Recent prototypical MIL approaches for pathology include AttnMISL~\cite{yao2020whole}, which aggregates patch embeddings within the same cluster, followed by aggregating the pooled cluster embeddings. Following recent advances in visual self-supervised learning, prototypes have been used for constructing unsupervised slide features via pooling similar patch embeddings into a concatenated representation (H2T~\cite{VU2023handcrafted}) or measuring the proportion of prototypes assignments in WSIs (HPL~\cite{quiros2023mapping}). We note that H2T and HPL have limitations in not encoding cardinality or not including deep visual features, which are relevant for interpretability and solving panoramic tasks. Moreover, as many pretrained vision encoders in CPath are pretrained on TCGA, prototypical patterns lack extensive evaluation of out-of-domain performance.
\begin{figure*}[t]
   \centering
   \includegraphics[width=1\linewidth]{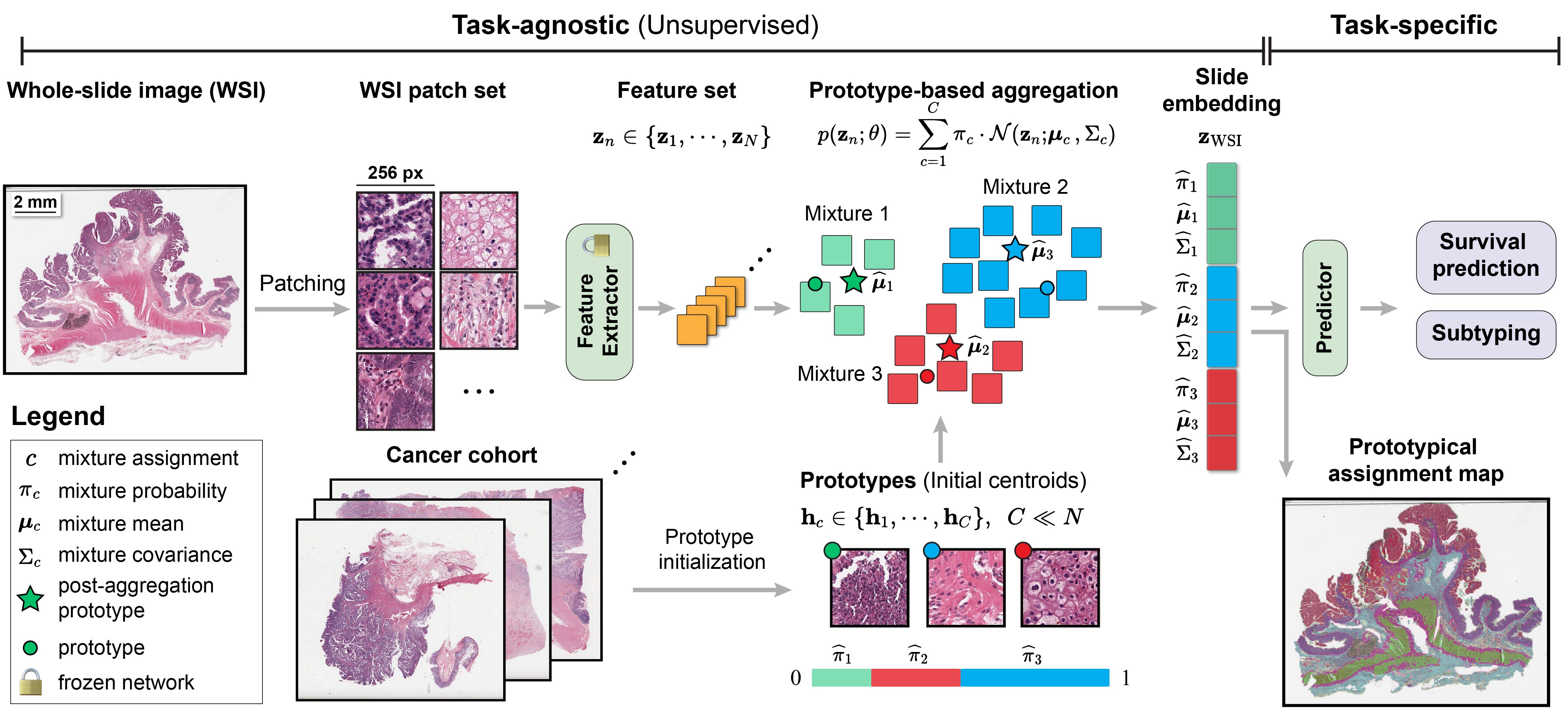}
   \caption{\textbf{Overview of $\ours$ workflow}. Whole-slide image (WSI) is segmented and patched into a set of WSI patches. A compressed feature for each patch is encoded through a feature extractor pretrained on a large histopathology dataset. $\ours$ uses the Gaussian mixture model for patch embedding distribution, with each mixture corresponding to a morphologically distinct prototype. The estimated model parameters are concatenated to form the slide representation, which can be used as input to a predictor module for clinical downstream tasks and visualized as a prototypical assignment map.
   }
   \label{fig:main}
\end{figure*}

\section{Methods}\label{sec:methods}
We present $\ours$, an unsupervised slide representation learning framework based on a compact set of prototypes with GMM (\textbf{Fig.}~\ref{fig:main}). We first explain the GMM setup and its connection to a slide embedding (Section~\ref{sec:aggregation}). We then present how it is used for downstream tasks (Section~\ref{sec:downstream}) and prototype-based interpretation (Section~\ref{sec:interpretability}).

\subsection{Prototype-based aggregation}\label{sec:aggregation}
Given a WSI for subject $j$, we tessellate it into small non-overlapping patches $\mathbf{X}^j=\{\x_1^j,\cdots, \x^j_{N_j}\}$ with $\x_n^j\in\mathbb{R}^{W\times H \times 3}$. We then employ a feature extractor $f_{\text{enc}}(\cdot)$ pretrained on large archives of histopathology images~\cite{chen2024towards}, to extract a representative and compressed embedding from each patch.
The set of extracted embeddings $\mathbf{Z}^j=\{\z_1^j, \cdots, \z_{N_j}^j \}$ with $\z_n^j = f_{\text{enc}}(\x_n^j) \in \mathbb{R}^d$ is then \textit{aggregated} to construct a slide embedding $\z_{\text{WSI}}^j$.

We aim to represent the set $\mathbf{Z}^j$ with a small set of prototypes $\largeh=\{\mathbf{h}_1,\cdots,\mathbf{h}_C\}$ with $\h_c\in\mathbb{R}^d$, $C \ll N_j$, without compromising essential morphological information.
Using the prototypes as \textit{references}, each patch is aggregated (or mapped) to the references and form $\slide\in\mathbb{R}^{C\cdot M}$~\cite{kim2022differentiable, mialon2021a}
\begin{equation}
    \slide=[\sum_{n=1}^{N_j}\phi^j(\z_n^j, \mathbf{h}_1),\cdots,\sum_{n=1}^{N_j}\phi^j(\z_n^j, \mathbf{h}_C)],
\end{equation}
where $\phi^j(\cdot, \cdot):\mathbb{R}^d\times\mathbb{R}^d\rightarrow\mathbb{R}^M$ is a function that maps a pair of patch embedding and reference, based on the similarity between the two, to a post-aggregation prototype embedding. In our work, the typical dimensions are $C=8\sim 32$ and $N_j=10,000\sim 20,000$. The $\slide$ dimension is fixed such that a variable-length set of $N_j$ features can always be represented in fixed-length. 
 
To define and estimate the mapping function $\phi^j$, we introduce a probabilistic framework for patch embedding distribution and assume each $\z_n^j$ is generated from a GMM,
\begin{equation}\label{eq:gmm}
\begin{split}
   p(\z_n^j ; \theta^j) &= \sum_{c=1}^C p(c_n^j=c; \theta^j)\cdot p(\z_n^j| c_n^j=c;\theta^j)\\
   &= \sum_{c=1}^C\pi_c^j \cdot \mathcal{N}(\z_n^j; \boldsymbol{\mu}_c^j, \Sigma_c^j),\,\, s.t.\sum_{c=1}^C\pi_c^j=1,\\ 
\end{split}
\end{equation}
where $\pi_c^j$ refers to the mixture probability of component $c$ in WSI set $j$ and $\theta^j$ refers to the set of parameters to be estimated, i.e., $\theta^j=\{\pi_c^j, \boldsymbol{\mu}_c^j, \Sigma_c^j \}_{c=1}^C$. For ease of computation, we use the diagonal covariance $\Sigma_c^j$. Intuitively, Eq.~\ref{eq:gmm} states that a morphological prototype and its variations correspond to a mixture component, with $\pi_c^j$ indicating the extent to which the pattern manifests in the $j^{\text{th}}$ WSI.

We formulate the slide embedding construction as that of estimating $\widehat{\theta}^j$ from $\mathbf{Z}^j$. To this end, we maximize the log-likelihood (or log-posterior if prior is introduced for $\theta^j$)
\begin{equation}
    \max_{\theta^j} \log p(\mathbf{Z}^j;\theta^j)=\max_{\theta^j}\sum_{n=1}^{N_j}\log p(\mathbf{z}^j_n;\theta^j).
\end{equation}

We use the expectation-maximization (EM) algorithm to obtain the maximum likelihood estimate $\theta^j$~\cite{dempster1977maximum, kim2022differentiable}, with derivation provided in \supp. The algorithm produces a posterior distribution $q(c_n^j=c|\z_n^j)$, which represents the probability that $\z_n^j$ is associated with prototype $c$.
At EM iteration $t+1$, it is given as
\begin{equation}\label{eq:posterior}
    q^{(t+1)}(c_n^j=c | \z_n^j) = \dfrac{\pi_c^{j, (t)}\mathcal{N}(\z_n^j; \boldsymbol{\mu}_c^{j,(t)}, \Sigma_c^{j,(t)})}{\sum_{c=1}^C\pi_c^{j,(t)}\mathcal{N}(\z_n^j; \boldsymbol{\mu}_c^{j,(t)}, \Sigma_c^{j,(t)})}.
\end{equation}
Eq.~(\ref{eq:posterior}) indicates that a patch is assigned to a certain prototype $c$ that is most similar, with the similarity measured in terms of weighted $\mathcal{L}_2$ distance. Moreover, the soft prototype assignment, i.e., $q^{(t+1)}(c_n^j=c | \z_n^j)>0,\,\forall c$, allows each patch to contribute towards all prototypes, in contrast to hard prototype assignment approaches~\cite{VU2023handcrafted, yu2023prototypical, quiros2023mapping, yao2020whole}. Using $q^{j,(t+1)}_{n,c}=q^{(t+1)}(c_n^j=c | \z_n^j)$ for notational simplicity, we can estimate $\theta^{j,(t+1)}$ as
\begin{equation}\label{eq:em}
\begin{split}
    \pi_c^{j, (t+1)} &= \frac{\sum_{n=1}^{N_j} q^{j,(t+1)}_{n,c}}{N_j},\,\,\, \boldsymbol{\mu}_c^{j,(t+1)}=\frac{\sum_{n=1}^{N_j} q^{j,(t+1)}_{n,c}\cdot\z_n^j}{\sum_{n=1}^{N_j} q^{j,(t+1)}_{n,c}}\\
    {\Sigma}_c^{j,(t+1)}&=\frac{\sum_{n=1}^{N_j} q^{j,(t+1)}_{n,c}\cdot (\z_n^j - \boldsymbol{\mu}_c^{j,(t+1)})^2}{\sum_{n=1}^{N_j} q^{j,(t+1)}_{n,c}}\\
\end{split}
\end{equation}
We set $\pi_c^{j,(0)}=1/C$, $\boldsymbol{\mu}_c^{j,(0)}=\h_c$, and $\Sigma_c^{j,(0)}=\mathbf{I}$, which serves as a morphology-aware initialization. 
Due to its iterative nature, EM can be placed as a neural network module. 
The initialization for $\{\mathbf{h}_c\}_{c=1}^C$ is performed with K-means clustering on the entire patch training set, constructed by aggregating patch embeddings from all training slides in a cohort.

Based on the final estimate $\widehat{\theta}^j$ after the EM convergence, the slide embedding $\z_{\text{WSI}}^j\in\mathbb{R}^{C\cdot M}$ with $M=1+2d$ can be represented as a concatenation of the elements in $\widehat{\theta}^j$, following set representation learning literature~\cite{mialon2021a, kim2022differentiable},  
\begin{equation}\label{eq:slide_embed}
    \begin{split}
    \z_{\text{WSI}}^j&=[\z_{\text{WSI},1}^j,\cdots, \z_{\text{WSI},C}^j]\\
    &=[\underbrace{\widehat{\pi}_1^{j}, \widehat{\boldsymbol{\mu}}_1^{j}, \widehat{\Sigma}_1^{j}}_{\sum_{n=1}^{N_j}\phi^j(\z_n^j, \mathbf{h}_1)}, \cdots, \underbrace{\widehat{\pi}_C^{j}, \widehat{\boldsymbol{\mu}}_C^{j}, \widehat{\Sigma}_C^{j}}_{\sum_{n=1}^{N_j}\phi^j(\z_n^j, \mathbf{h}_C)}].
    \end{split}
\end{equation}
We emphasize that while the prototypes $\{\h_c \}_{c=1}^C$ are shared across different WSIs, the parameter estimation and the slide embedding construction are performed per WSI.
Overall, the embedding $\slide$ satisfies two essential principles for a faithful WSI representation and good downstream performance. First, it accounts for the cardinality of each prototype explicitly through $\widehat{\pi}_c^{j}$ and implicitly through $\widehat{\boldsymbol{\mu}}_c^{j}$ and $\widehat{\Sigma}_c^{j}$. In addition, by concatenating the features (rather than averaging), feature vector for each morphological prototype is directly accessible for downstream tasks. 

\subsubsection{Connection to optimal transport}
The aggregation in $\ours$ can be seen as matching the empirical distribution of patch embeddings and the prototypes, defined as $\hat{p}_j$ and $\hat{q}_j$ respectively. Specifically, we have $\hat{p}_j = \sum_{n=1}^{N_j}a_n^j\cdot\delta_{\z_n^j}\quad\text{and}\quad\hat{q}_j = \sum_{c=1}^{C}\pi_c^j \cdot \delta_{\mathbf{h}_c},$
where $\sum_{n=1}^{N_j}a_n^j=\sum_{c=1}^C \pi_c^j=1$. OT aggregates $\{\z_n^j\}_{n=1}^{N_j}$ to $\{\mathbf{h}_c\}_{c=1}^C$ by minimizing the Wasserstein distance between $\hat{p}_j$ and $\hat{q}_j$, typically assuming uniform distribution for $\{a_n^j\}_{n=1}^{N_j}$ and $\{\pi_c^j\}_{c=1}^{C}$, i.e., $a_n^j=1/N_j$ and $\pi_c^j=1/C$~\cite{cuturi2013sinkhorn, mialon2021a}. While $\ours$ also assumes $a_n^j=1/N_j$, the mixture probability $\pi_c^j$ is estimated (Eq.~\ref{eq:em}). 
Furthermore, the OT solution can be seen as a special case of GMM solution with uniform prototype distribution~\cite{kim2022differentiable}.

\subsection{Downstream evaluation}\label{sec:downstream}

The embedding $\slide$ can be used as the input to a \textit{predictor} module $g(\cdot)$ for various downstream tasks. The predictor $g(\cdot)$ can be a linear layer (linear probing), or implemented as a multilayer perceptron (MLP). Instead of using the entire $\slide$ as an input to a MLP, we propose a structured MLP
\begin{equation}\label{eq:MLP}
    \begin{split}        
        \z_{\text{WSI}}'&=[g^{\text{indiv.}}_1(\z_{\text{WSI},1}),\ldots, g^{\text{indiv.}}_C(\z_{\text{WSI},C})]\\
        g(\z_{\text{WSI}})&=g^{\text{pred.}}(\z_{\text{WSI}}'),\\
    \end{split}
\end{equation}
where $j$ is dropped for notational simplicity, and $g^{\text{pred.}}$ and $\{g^{\text{indiv.}}_c\}_{c=1}^C$ assume one of $\{\text{Identity}, \text{Linear}, \text{MLP}\}$. 
Eq.~\ref{eq:MLP} leverages that $\slide$, which is a concatenation of mixture estimates, can be decomposed along each prototype and learns per-prototype nonlinear mapping $g^{\text{indiv.}}_c$. This is not possible with a typical MIL: First, the large and variable size of $N$ prohibits the learning of $\{g_n^{\text{indiv.}}\}_{n=1}^{N}$. Moreover, the permutation invariance makes finding an ``appropriate'' function $g_n^{\text{indiv.}}$ for a given patch embedding $\z_n$ non-trivial. 

\subsection{Interpretability}\label{sec:interpretability}
Based on the estimated prototype assignment probability $q$, we propose two approaches for interpretability. First, for each patch embedding $\z_n^j$, we assign the prototype with the highest posterior probability,
\begin{equation}
    c_n^j = \operatorname{argmax}_{c} q(c_n^j=c \mid \z_n^j),
\end{equation}
and overlay the prototype assignments onto WSI, visualizing how pathology visual concepts are distributed within each WSI (prototypical assignment map, \textbf{Fig.}~\ref{fig:heatmap}), with $\widehat{\pi}_c^j$ quantifying the extent of the distribution. For a specific prototype $c'$, we can also visualize how morphologically similar each patch embedding is to the prototype using $q(c_n^j=c' \mid \z_n^j)$ (\textbf{Fig.}~\ref{fig:top}). 
\section{Experiments}\label{sec:experiments}
\subsection{Datasets}
\textbf{Subtyping} 
We evaluate $\ours$ on four different subtyping tasks: EBRAINS fine subtyping (30 classes) and coarse subtyping (12 classes) for rare brain cancer types~\cite{gatta2017burden, roetzer2022digital}, Non-Small Cell Lung Carcinoma (NSCLC) subtyping on TCGA and CPTAC (2 classes), and ISUP grading based on Prostate cancer grade assessment (PANDA) challenge (6 classes)~\cite{bulten2020automated, bulten2022artificial}. We use balanced accuracy and weighted F1 metrics for evaluation on EBRAINS and NSCLC subtyping task and Cohen's $\kappa$ for the ISUP grading task. 

\noindent\textbf{Survival} We evaluate $\ours$ on TCGA across several cancer types: Breast
Invasive Carcinoma (BRCA), Colon and Rectum Adenocarcinoma (CRC), Bladder Urothelial Carcinoma (BLCA), Uterine corpus endometrial carcinoma (UCEC), Kidney renal clear cell carcinoma (KIRC), and Lung adenocarcinoma (LUAD). For TCGA, we use the 5-fold site-stratified cross-validation. For cancer types with external validation datasets (\textbf{KIRC}: CPTAC, \textbf{LUAD}: CPTAC, NLST), we use the models trained on TCGA and evaluate on the external dataset.
We use the concordance index (c-index) for evaluation. To address the shortcomings of overall survival accounting for non-cancerous deaths~\cite{liu2018integrated, carmichael2022incorporating, jaume2023modeling}, we use disease-specific survival (DSS). Additional details can be found in $\supp$.

\begin{table*}[!ht]
\centering
\small
\caption{\textbf{Subtyping prediction} Results of $\ours$ and baselines for four different subtyping tasks. All methods use UNI features~\cite{chen2024towards}. Best performance in \textbf{bold}, second best \underline{underlined}. AttnMISL, ProtoCounts, H2T, OT, and $\ours$ use $C=16$ prototypes.}
\begin{tabular}{ll|cc|cc|cc|cc}
\toprule
&\textbf{Train on} & \multicolumn{2}{c}{EBRAINS} & \multicolumn{2}{c}{EBRAINS} & \multicolumn{2}{c}{TCGA-NSCLC} & \multicolumn{2}{c}{PANDA} \\
& & \multicolumn{2}{c}{(fine, 30 classes)} & \multicolumn{2}{c}{(coarse, 12 classes)} & \multicolumn{2}{c}{(2 classes)} & \multicolumn{2}{c}{(6 classes)} \\
\cline{3-10}
&\textbf{Test on} & \multicolumn{2}{c|}{EBRAINS} & \multicolumn{2}{c|}{EBRAINS} & TCGA & CPTAC & Karolinska & Radboud \\
&& (Bal. acc.) & (F1) & (Bal. acc.) & (F1) & (Bal. acc.) & (Bal. acc.) & (Cohen's $\kappa$) & (Cohen's $\kappa$) \\
\midrule
\parbox[t]{0mm}{\multirow{5}{*}{\rotatebox[origin=c]{90}{{\textbf{Supervised.}}}}} 
&ABMIL~\cite{ilse2018attention} & 0.674 & 0.744 & 0.834 & 0.906 & 0.949 & \underline{0.904} & \underline{0.935} & 0.918 \\
&TransMIL~\cite{shao2021transmil} & \textbf{0.701} & \underline{0.758} & \underline{0.848} & \textbf{0.921} & 0.959 & 0.867 & \textbf{0.942} & \underline{0.922} \\
&DSMIL~\cite{li2021dual} & 0.648 & 0.698 & 0.824 & 0.882 & \textbf{0.980} & 0.791 & 0.909 & 0.911\\
&AttnMISL~\cite{yao2020whole} & 0.534 & 0.636 & 0.647 & 0.823 & 0.888 & 0.823 & 0.882 & 0.894 \\
&ILRA~\cite{xiang2023exploring} & 0.618 & 0.695 & 0.820 & 0.896 & 0.939 & 0.887 & 0.931 & 0.925 \\
\midrule
\parbox[t]{0mm}{\multirow{4}{*}{\rotatebox[origin=c]{90}{{\textbf{Unsup.}}}}} 
&DeepSets~\cite{zaheer2017deep} & 0.033 & 0.073 & 0.082 & 0.2 & 0.571 & 0.707 & $<$ 0 & $<$ 0 \\
&ProtoCounts~\cite{yu2023prototypical, quiros2023mapping} & 0.038 & 0.018 & 0.097 & 0.079 & 0.429 & 0.569 & $<$ 0 & 0.13 \\
&H2T~\cite{VU2023handcrafted} & 0.117 & 0.223 & 0.181 & 0.421 & 0.929 & 0.821 & 0.457 & 0.755  \\
&OT~\cite{mialon2021a} & \underline{0.700} & 0.756 & 0.837 & \underline{0.915} & 0.950 & 0.867 & 0.817 & 0.883  \\
\midrule
\parbox[t]{0mm}{\multirow{5}{*}{\rotatebox[origin=c]{90}{{\textbf{Ours}}}}} 
&$\ours_{\text{Top}}$ + lin. & \multirow{1}{*}{0.471} & \multirow{1}{*}{0.571} & \multirow{1}{*}{0.554} & \multirow{1}{*}{0.758} & \multirow{1}{*}{0.857} & \multirow{1}{*}{0.833} & \multirow{1}{*}{0.631} & \multirow{1}{*}{0.689} \\
&$\ours_{\text{Bot.}}$ + lin. & \multirow{1}{*}{0.038} & \multirow{1}{*}{0.080} & \multirow{1}{*}{0.138} & \multirow{1}{*}{0.329} & \multirow{1}{*}{0.602} & \multirow{1}{*}{0.705} & \multirow{1}{*}{0.071} & \multirow{1}{*}{0.000} \\
&$\ours_{\text{WA}}$ + lin. & \multirow{1}{*}{0.497} & \multirow{1}{*}{0.598} & \multirow{1}{*}{0.569} & \multirow{1}{*}{0.784} & \multirow{1}{*}{0.888} & \multirow{1}{*}{0.860} & \multirow{1}{*}{0.663} & \multirow{1}{*}{0.787} \\
&\textbf{$\ours_{\text{All}}$ + lin.} & \multirow{1}{*}{0.691} & \multirow{1}{*}{0.756} & \multirow{1}{*}{0.829} & \multirow{1}{*}{0.904} & \multirow{1}{*}{0.939} & \multirow{1}{*}{0.882} & \multirow{1}{*}{0.866} & \multirow{1}{*}{0.909} \\
&\textbf{$\ours_{\text{All}}$ + MLP} & \multirow{1}{*}{0.693} & \multirow{1}{*}{\textbf{0.760}} & \multirow{1}{*}{\textbf{0.854}} & \multirow{1}{*}{0.908} & \multirow{1}{*}{\textbf{0.980}} & \multirow{1}{*}{\textbf{0.906}} & \multirow{1}{*}{0.923} & \multirow{1}{*}{\textbf{0.931}}  \\
\bottomrule
\end{tabular}
\label{tab:clf_main}
\end{table*}

\subsection{Baselines}
We employ 1) \textit{unsupervised} baselines, which use unsupervised slide representation followed by the task-specific linear network, and 2) \textit{supervised} baselines, which construct supervised slide representation for each task. For the \textit{unsupervised} baselines, we use the following: 1) \textbf{DeepSets}~\cite{zaheer2017deep} The slide embedding $\slide\in\mathbb{R}^d$ is formed by averaging all the features in the set. 2) \textbf{ProtoCounts}~\cite{quiros2023mapping, yu2023prototypical} K-means clustering is performed on the cohort-aggregated set of features. The slide embedding $\slide\in\mathbb{R}^C$ is a count vector of the number of patches assigned to each cluster. 3) \textbf{H2T}~\cite{VU2023handcrafted} The patch embeddings are clustered and averaged within each cluster. The averaged cluster centroids are concatenated, with $\slide\in\mathbb{R}^{C\cdot d}$. 4) \textbf{Optimal Transport (OT)}~\cite{mialon2021a} The patch features of a WSI is aggregated to a set of prototypes with OT~\cite{cuturi2013sinkhorn}, with $\slide\in\mathbb{R}^{C\cdot d}$. OT assumes uniform mixture probability, i.e., $\pi_c^j=1/C, \forall c$.

We also implement the following supervised baselines: Attention-based MIL (ABMIL)~\cite{ilse2018attention}, Transformer-based MIL (TransMIL)~\cite{shao2021transmil}, Prototype-clustering based MIL (AttnMISL)~\cite{yao2020whole}, and low-rank MIL (ILRA)~\cite{xiang2023exploring}.

For $\ours$, we experiment with variations of $\slide$ to better understand our model: 1) \textbf{All} (original): All mixture parameters are concatenated, $\slide\in\mathbb{R}^{C(1+2d)}$. 2) \textbf{Weighted avg.} (WA): $\boldsymbol{\mu}_c$ and $\Sigma_c$ weighted-averaged by $\pi_c$ and concatenated, $\slide\in\mathbb{R}^{2d}$. 3) \textbf{Top (Bottom)}: Parameters for mixture component with the largest (smallest) $\widehat{\pi}^j_c$ is selected, $\slide\in\mathbb{R}^{(1+2d)}$. We use either linear (+lin.) or nonlinear head (+MLP) on top of $\slide$.

For the feature extractor $f_{\text{enc}}(\cdot)$, which is the same for all baselines used in this work, we used UNI~\cite{chen2024towards}, a ViT-L/16 DINOv2~\cite{dosovitskiy2021image,oquab2023dinov2} that was pre-trained on a large internal histology dataset of $1\times10^8$ patches from $1\times 10^6$ WSIs\footnote{Accessible at: https://github.com/mahmoodlab/UNI}. We also experiment with other feature encoders, CTransPath~\cite{wang2022transformer} and ResNet50~\cite{he2016deep}, the results of which can be found in $\supp$.

\subsection{Implementation}
WSIs at $20\times$ magnification ($0.5\, \mu m$/pixel) are patched with non-overlapping patches of $256\times 256$ pixels. For each WSI, we use all patches without sampling. We found that a single EM step is sufficient for convergence. The prototypes $\mathbf{H}$ are constructed from K-means clustering on the set of patches aggregated from the training cohort (all slides) for each task. The same $\mathbf{H}$ is used for AttnMISL, ProtoCounts, H2T, OT, and $\ours$. Additional details on training and loss functions can be found in $\supp$.

\section{Results}\label{sec:results}

\subsection{Subtyping and survival prediction}
Subtyping and survival prediction results are shown in Table~\ref{tab:clf_main} and Table~\ref{tab:survival_main}. Overall, $\ours$ consistently outperforms or is on par with all supervised and unsupervised baselines. We highlight key insights and provide hypotheses for the high performance of $\ours$.

\begin{table*}[!ht]
\centering
\small
\caption{\textbf{Survival prediction} Results of $\ours$ and baselines for measuring patient disease-specific survival based on c-index. All methods use UNI features~\cite{chen2024towards}. Best performance in \textbf{bold}, second best \underline{underlined}. All models and prototypes are trained on TCGA. AttnMISL, ProtoCounts, H2T, OT, and $\ours$ use $C=16$ prototypes. Standard deviation (in parentheses) are reported over five runs.}
\begin{tabular}{ll|c|c|c|c|cc|ccc}
\toprule
&\textbf{Dataset} & BRCA & CRC & BLCA  & UCEC & \multicolumn{2}{c|}{KIRC} & \multicolumn{3}{c}{LUAD} \\
\cline{3-11}
&\textbf{Test on} &TCGA & TCGA & TCGA & TCGA & TCGA & CPTAC & TCGA & CPTAC & NLST \\
\midrule
\parbox[t]{0mm}{\multirow{10}{*}{\rotatebox[origin=c]{90}{{\textbf{Supervised}}}}}
&\multirow{2}{*}{ABMIL~\cite{ilse2018attention}}     & 0.644 & 0.608 & 0.550 & 0.669 & 0.684 & 0.613 & 0.654 & 0.572 & 0.519 \\
&& ($\pm$ 0.05) & ($\pm$ 0.09) & ($\pm$ 0.06) & ($\pm$ 0.07)  & ($\pm$ 0.06) & ($\pm$ 0.06) & ($\pm$ 0.06) & ($\pm$ 0.03) & ($\pm$ 0.04) \\
&\multirow{2}{*}{TransMIL~\cite{shao2021transmil}}  & 0.612 & \textbf{0.684} & 0.595 &  0.695 & 0.671 & 0.639 & 0.665 & 0.555 & 0.484 \\
&& ($\pm$ 0.07) & ($\pm$ 0.06) & ($\pm$ 0.06) & ($\pm$ 0.08) & ($\pm$ 0.10) & ($\pm$ 0.04) & ($\pm$ 0.10) & ($\pm$ 0.03) & ($\pm$ 0.05) \\
&\multirow{2}{*}{DSMIL~\cite{li2021dual}} & 0.496 & 0.5 & 0.501 & 0.497 & 0.5 & 0.5 & 0.501 & 0.502 & 0.5\\
&& ($\pm$ 0.00) & ($\pm$ 0.00) & ($\pm$ 0.00) & ($\pm$ 0.00) & ($\pm$ 0.00) & ($\pm$ 0.00) & ($\pm$ 0.00) & ($\pm$ 0.00) & ($\pm$ 0.00) \\
&\multirow{2}{*}{AttnMISL~\cite{yao2020whole}} & 0.627  & 0.639 & 0.485 & 0.581 & 0.649 & 0.608 & 0.673 & 0.632 & 0.577\\
&& ($\pm$ 0.08) & ($\pm$ 0.10) & ($\pm$ 0.06) & ($\pm$ 0.12) & ($\pm$ 0.09) & ($\pm$ 0.06) & ($\pm$ 0.10) & ($\pm$ 0.03) & ($\pm$ 0.04)\\
&\multirow{2}{*}{ILRA~\cite{xiang2023exploring}} & 0.649 & 0.555 & 0.550 & 0.632 & 0.637 & 0.611 & 0.586 & \underline{0.651} & 0.482 \\
&& ($\pm$ 0.10) & ($\pm$ 0.10) & ($\pm$ 0.04) & ($\pm$ 0.02) & ($\pm$ 0.14) & ($\pm$ 0.03) & ($\pm$ 0.06) & ($\pm$ 0.05) & ($\pm$ 0.01)\\
\midrule
\parbox[t]{0mm}{\multirow{8}{*}{\rotatebox[origin=c]{90}{{\textbf{Unsupervised}}}}}
&\multirow{2}{*}{DeepSets~\cite{zaheer2017deep}}  & 0.673  & 0.563 & 0.581 & 0.730 & 0.715 & 0.634 & 0.652 & 0.550 & 0.509 \\
&& ($\pm$ 0.11) & ($\pm$ 0.10) & ($\pm$ 0.05) & ($\pm$ 0.05) & ($\pm$ 0.08) & ($\pm$ 0.01) & ($\pm$ 0.05) & ($\pm$ 0.01) & ($\pm$ 0.04) \\
&\multirow{2}{*}{ProtoCounts~\cite{yu2023prototypical, quiros2023mapping}} & 0.490 & 0.552 & 0.533 & 0.441 & 0.461 & 0.503 & 0.460 & 0.577 & 0.500\\
&& ($\pm$ 0.11) & ($\pm$ 0.06) & ($\pm$ 0.09) & ($\pm$ 0.06) & ($\pm$ 0.06) & ($\pm$ 0.09) & ($\pm$ 0.11) & ($\pm$ 0.11) & ($\pm$ 0.01) \\
&\multirow{2}{*}{H2T~\cite{VU2023handcrafted}}  & 0.672  & 0.639 & 0.566 & 0.715 & 0.703 & 0.631 & 0.662 & 0.583  & 0.603 \\
&& ($\pm$ 0.07) & ($\pm$ 0.11) & ($\pm$ 0.05) & ($\pm$ 0.09) & ($\pm$ 0.11) & ($\pm$ 0.04) & ($\pm$ 0.09) & ($\pm$ 0.03) & ($\pm$ 0.04) \\
&\multirow{2}{*}{OT~\cite{mialon2021a}}  & \underline{0.755}  & 0.622 & \underline{0.603} & 0.747 & 0.695 & 0.650 & \textbf{0.687} & 0.641  & 0.495 \\
&& ($\pm$ 0.06) & ($\pm$ 0.09) & ($\pm$ 0.04) & ($\pm$ 0.08) & ($\pm$ 0.09) & ($\pm$ 0.02) & ($\pm$ 0.08) & ($\pm$ 0.02) & ($\pm$ 0.04) \\
\midrule
\parbox[t]{0mm}{\multirow{10}{*}{\rotatebox[origin=c]{90}{{\textbf{Ours}}}}} 
&\multirow{2}{*}{$\ours_{\text{Top}}$ + lin.} & 0.718 & 0.534 & 0.543 & 0.707 & \textbf{0.741} & 0.608 & 0.575 & 0.607 & 0.417\\
&& ($\pm$ 0.11) & ($\pm$ 0.08) & ($\pm$ 0.05) & ($\pm 0.08$) &  ($\pm$ 0.11) & ($\pm$ 0.01) & ($\pm$ 0.05) & ($\pm$ 0.02) & ($\pm$ 0.06)\\
&\multirow{2}{*}{$\ours_{\text{Bot.}}$ + lin.}  & 0.578 & 0.452 & 0.494 & 0.570 & 0.524 & 0.532 & 0.592 & 0.539  & 0.519 \\
 && ($\pm$ 0.13) & ($\pm$ 0.08) & ($\pm$ 0.08) & ($\pm$ 0.11) &  ($\pm$ 0.14) & ($\pm$ 0.08) & ($\pm$ 0.10) & ($\pm$ 0.12) & ($\pm$ 0.02)\\
&\multirow{2}{*}{$\ours_{\text{WA}}$ + lin.} & 0.670 & 0.647 & 0.586 & \underline{0.753} & \underline{0.730} & 0.623 & 0.654 & 0.461 & 0.482 \\
& & ($\pm$ 0.09) & ($\pm$ 0.12) & ($\pm$ 0.04) & ($\pm$ 0.09) & ($\pm$ 0.07) & ($\pm$ 0.01) & ($\pm$ 0.07) & ($\pm$ 0.01) & ($\pm$ 0.06)\\
&\multirow{2}{*}{\textbf{$\ours_{\text{All}}$ + lin.}} & 0.722 & 0.645 & 0.602 & 0.751 & 0.703 & \underline{0.649} & 0.672 & 0.568 & \underline{0.623} \\
&& ($\pm$ 0.07) & ($\pm$ 0.07) &  ($\pm$ 0.05) &  ($\pm$ 0.11) & ($\pm$ 0.13) & ($\pm$ 0.04) & ($\pm$ 0.06) & ($\pm$ 0.05) & ($\pm$ 0.07)\\
&\multirow{2}{*}{\textbf{$\ours_{\text{All}}$ + MLP}} & \textbf{0.758} & \underline{0.665} & \textbf{0.612} & \textbf{0.757} & 0.716 & \textbf{0.691} & \underline{0.685}& \textbf{0.653} & \textbf{0.634}  \\
& & ($\pm$ 0.06) & ($\pm$ 0.10) & ($\pm$ 0.07) & ($\pm$ 0.10) & ($\pm$ 0.10) & ($\pm$ 0.03) & ($\pm$ 0.06) & ($\pm$ 0.04) & ($\pm$ 0.04) \\
\bottomrule
\end{tabular}
\label{tab:survival_main}
\end{table*}

\noindent\textbf{$\ours$ \emph{vs.} supervised MIL} $\ours_{\text{All}}$+MLP outperforms or is on par with the best-performing supervised baseline on subtyping (TransMIL) and survival prediction tasks (mix). With linear probing, $\ours_{\text{All}}$+lin. remains competitive against MIL on subtyping and performs better on most cancer types in survival prediction, demonstrating the strong representation quality of $\slide$. This is encouraging as $\ours$ builds a slide representation in an unsupervised fashion, unlike MIL which learns a patch aggregation end-to-end with the downstream tasks. 
Interestingly, despite relying on a similar prototype construction as $\ours$, AttnMISL~\cite{yao2020whole} performs consistently lower than $\ours$. We attribute this difference to AttnMISL \emph{averaging} the prototypes with attention weights, whereas $\ours$ builds a slide embedding by \emph{concatenating} them. Our baseline $\ours_{\text{WA}}$+lin. further confirms this by showing that averaging leads to a consistently lower performance.

\noindent\textbf{$\ours$ vs. unsupervised baselines} We observe that $\ours_{\text{All}}$+MLP outperforms most unsupervised baselines on subtyping and survival prediction tasks.
We attribute this gain to two design principles behind $\ours$: (1) prototypes are represented as low-dimensional feature vectors, and (2) the resulting slide embedding encodes the cardinality of each prototype, \emph{i.e.,} their extent in the WSI. In comparison, ProtoCounts only encodes the count information, leading to poor performance. Interestingly, DeepSets, which builds slide embeddings as the sum of all patch embeddings, and similarly our baseline $\ours_{\text{WA}}$+lin., which takes weighted averaging of the prototype features, lead to poor subtyping performance despite implicitly encoding both deep patch representations and cardinality. We hypothesize that subtyping requires a mechanism to ``isolate" discriminative information, which can be implemented using attention (as in ABMIL and TransMIL), or using prototype concatenation as in $\ours$. 

Unsurprisingly, H2T and OT come closest to $\ours$ on subtyping tasks as they also aggregate the patches to the prototypes (albeit with different mechanisms from $\ours$) and use concatenation. However, on ISUP grading in prostate cancer which is clinically assessed using the primary and secondary Gleason patterns, $\ours$ sees significant performance increases. Lastly, H2T and OT lack explicit mechanisms to incorporate cardinality into slide representation, an important feature of $\ours$ for interpretability (Section~\ref{sec:interpretability_result}).

In this context, $\ours$ appears as a comprehensive unsupervised slide representation method that concatenates deep prototype representations along with their cardinality.

\noindent\textbf{$\ours$ ablations} We further ablate $\ours$ by retaining only a single component with the highest (lowest) $\widehat{\pi}_c$, \emph{i.e.,} $\ours_{\text{Top}}$ ($\ours_{\text{Bot.}}$). We observe that both $\ours_{\text{Top}}$ and $\ours_{\text{Bot.}}$ performs consistently poorly. This reaffirms that capturing morphological heterogeneity is crucial for accurate prediction of a patient's clinical outcome~\cite{marusyk2020intratumor, vitale2021intratumoral}. That $\ours_{\text{Bot.}}$ is the lowest-performing agrees with our intuition, as subtypes and grades are most often determined by pathologists based on visual cues that must be integrated across the entirety of the tumor, rather than utilizing only a particular region or morphology within the whole. Interestingly, $\ours_{\text{Top}}$ performs relatively well for the NSCLC subtyping task, which we attribute to it being a relatively simple binary classification task and the most populous component (highest $\widehat{\pi}_c$) for the NSCLC WSIs on average being tumor.

\noindent\textbf{$\ours$ with linear vs. non-linear head}
In all tasks, $\ours_{\text{All}}$+MLP consistently boosts the performance over $\ours_{\text{All}}$+lin., demonstrating additional predictive capability enabled by per-prototype non-linearity modeling.  

\subsection{Interpretability}\label{sec:interpretability_result}
To understand which prototypes are used in learning slide representations, we visualize (1) the patch-level prototype assignments per WSI via probability $q(c_n^j=c \mid \z_n^j)$, and (2) the distribution of mixture components $\widehat{\pi}^j_c$ within and across all WSIs in the cohort. 

\begin{figure*}[t]
   \centering
   \includegraphics[width=1\linewidth]{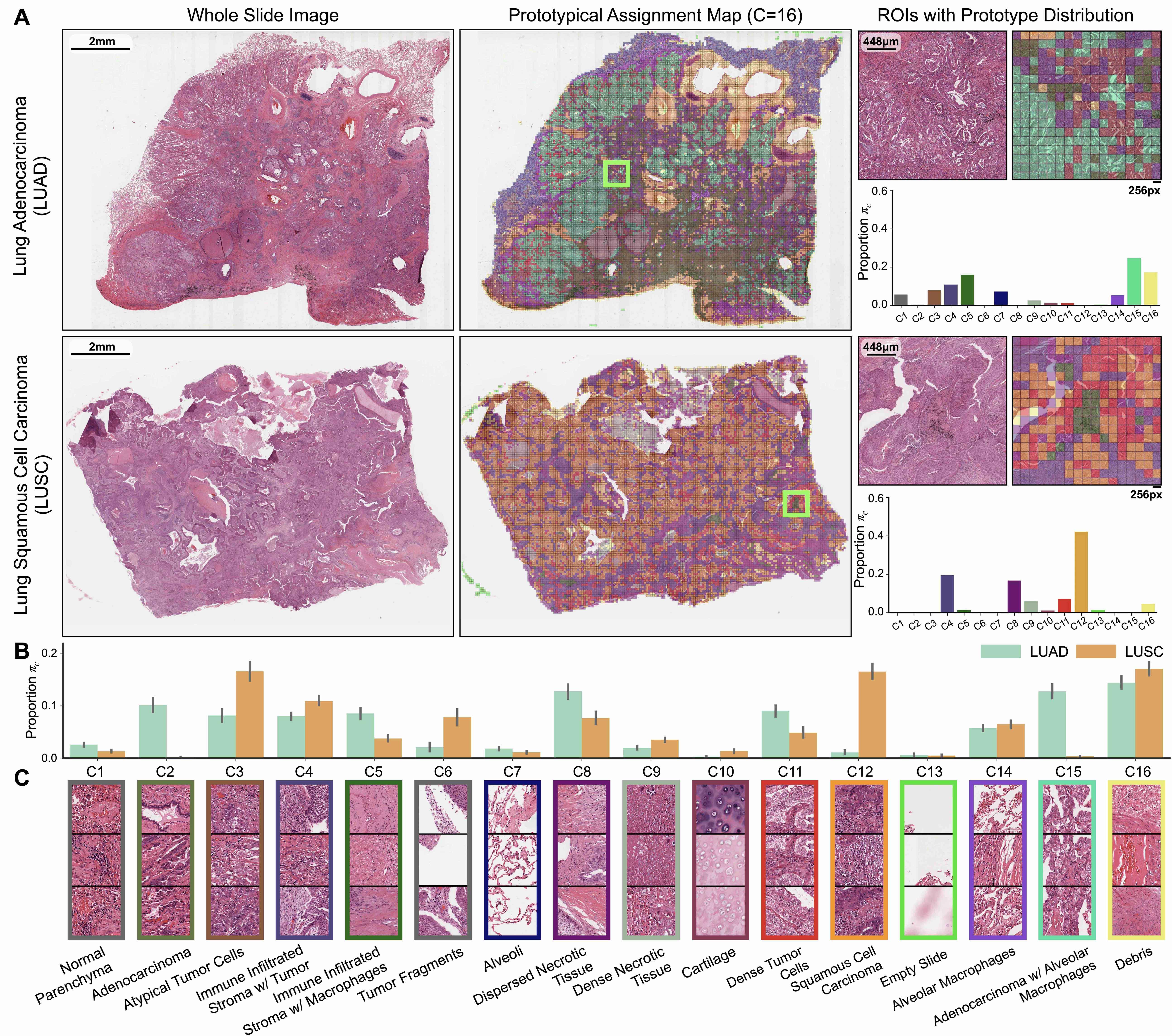}
   \caption{\textbf{Prototype-oriented heatmap interpretation}. (A) Examples of WSIs and prototypical assignment maps from LUAD and LUSC, with estimated prototype distribution $\hat{\pi}_c$ for each WSI. (B) Prototype distribution and morphological annotations by a board-certified pathologist in the NSCLC cohort. The adenocarcinoma prototypes (C2, C15) and squamous cell carcinoma (C12) appear exclusively in LUAD and LUSC respectively, showing that $\ours$ can correctly capture essential morphological cues in the tissue. 
   }
   \label{fig:heatmap}
\end{figure*}

Overall, we find that GMMs are a simple yet powerful framework for mapping the spatial organization of histologic visual concepts in the tissue microenvironment. Visual assessment by a board-certified pathologist (D.F.K.W.) revealed that prototypical patterns reflect distinct morphological phenotypes of tumor-, tumor-associated stromal, and immune-cell populations as well as normal tissue components (\textbf{Fig.}~\ref{fig:heatmap}C). In NSCLC, we discover prototypes that correspond to adenocarcinoma (turquoise, C2 and C15) and squamous cell carcinoma (orange, C12) patterns (\textbf{Fig.}~\ref{fig:heatmap}A,B). Analysis of $\widehat{\pi}_c$ shows that these patterns almost exclusively appeared in LUAD and LUSC slides (\textbf{Fig.}~\ref{fig:heatmap}C). In CRC, the distribution of prototypical patterns had strong concordance with existing tissue annotations for CRC tissue types (CRC-100K)~\cite{kather2019predicting}, with further visualizations presented in the \supp.

\subsection{Further ablations}
We run further ablation studies and present additional insights in $\supp$. Overall, $\ours$ is robust across different choices of the number of prototypes $C$, survival loss functions, and feature encoders.
\section{Conclusion and limitations}\label{sec:conclusion}
We present $\ours$, a prototype-based aggregation framework for learning unsupervised slide representations with Gaussian mixtures as the patch distribution. We believe this is an important addition to the emerging group of slide representation studies, with the unsupervised nature of $\ours$ making it readily applicable to diverse tasks. Limitations include using $C=16$ for all tasks, which may lead to over- or under-clustering for certain cancers. Future work includes introducing more expressive mixture models for patch distributions, determining the number of prototypes in a data-driven manner, and evaluating on rare cancer cohorts with small sample sizes.

\clearpage
{
    \small
    \bibliographystyle{ieeenat_fullname}
    \bibliography{main}
}

\clearpage
\setcounter{page}{1}
\setcounter{section}{0}
\setcounter{figure}{0}
\setcounter{table}{0}
\setcounter{equation}{0}

\renewcommand{\figurename}{Figure}
\renewcommand{\thefigure}{S\arabic{figure}}
\renewcommand{\tablename}{Table}
\renewcommand{\thetable}{S\arabic{table}}

\maketitlesupplementary


\section{Mathematical derivations of EM algorithm}
Given the Gaussian mixture model (GMM) for a generative model for each individual patch embedding, we provide a detailed derivation for estimation of 1) the posterior probability for the prototype assignment $q(c|\z_n^j;\theta_j)$ and 2) the GMM parameters $\theta^j=\{\pi_c^j, \boldsymbol{\mu}_c^j, \Sigma_c^j \}$. Given the GMM specification,

\begin{equation}\label{eq:gmm_supp}
\begin{split}
   p(\z_n^j ; \theta^j) &= \sum_{c=1}^C p(c_n^j=c; \theta^j)\cdot p(\z_n^j| c_n^j=c;\theta^j)\\
   &= \sum_{c=1}^C\pi_c^j \cdot \mathcal{N}(\z_n^j; \boldsymbol{\mu}_c^j, \Sigma_c^j),\,\, s.t.\sum_{c=1}^C\pi_c^j=1,\\ 
\end{split}
\end{equation}
the goal is to estimate $\theta^j$ that maximizes the log-likelihood
\begin{equation}
    \max_{\theta^j} \log p(\mathbf{Z}^j;\theta^j)=\max_{\theta^j}\sum_{n=1}^{N_j}\log p(\mathbf{z}^j_n;\theta^j).
\end{equation}
Using Jensen's inequality, we obtain the following lower bound for the log-likelihood,
\begin{equation}
    \begin{split}
        &\sum_{n=1}^{N_j}\log p(\mathbf{z}^j_n;\theta^j) \\
        &=\sum_{n=1}^{N_j}\log \sum_{c=1}^C p(\z_n^j, c_n^j=c;\theta^j)\\
        &=\sum_{n=1}^{N_j}\log \sum_{c=1}^C q(c_n^j=c|\z_n^j;\theta^j_{\text{old}})\cdot\frac{p(\z_n^j, c_n^j=c;\theta^j)}{q(c_n^j=c|\z_n^j;\theta^j_{\text{old}})}\\
        &\geq \sum_{n=1}^{N_j}\sum_{c=1}^C q(c_n^j=c|\z_n^j;\theta^j_{\text{old}}) \log \frac{p(\z_n^j, c_n^j=c;\theta^j)}{q(c_n^j=c|\z_n^j;\theta^j_{\text{old}})}\\
        &= \sum_{n=1}^{N_j}\underbrace{E_{q(c_n^j=c|\z_n^j;\theta^j_{\text{old}})}\left[\log p(\z_n^j, c_n^j=c;\theta^j)\right]}_{Q(\theta^j;\theta^j_{\text{old}})}\\
        &\,\,\quad -\sum_{n=1}^{N_j}\underbrace{E_{q(c_n^j=c|\z_n^j;\theta^j_{\text{old}})}\left[q(c_n^j=c|\z_n^j;\theta^j_{\text{old}})\right]}_{-H(C;\theta^j_{\text{old}})}.\\
    \end{split}
\end{equation}
This allows us to substitute the problem of maximizing the log-likelihood with that of maximizing a surrogate function, which in our case is the lower bound given by Jensen's inequality. It can be shown that increasing the lower bound with respect to $\theta^j$ leads to monotonically increasing log-likelihood~\cite{dempster1977maximum, kim2022differentiable}. The optimization of the surrogate function towards maximizing the log-likelihood is often referred to as the Expectation-Maximization (EM) algorithm, which iteratively alternates between the E-step and the M-step. 

The surrogate function is comprised of the two terms, $Q(\theta^j;\theta^j_{\text{old}})$ and $H(C;\theta^j_{\text{old}})$, both of which are expectations with respect to the posterior probability of prototype assignment, \emph{i.e.,} $q(c_n^j=c|\z_n^j;\theta^j_{\text{old}})$. In the E-step, we use Bayes' rule to compute the posterior probability and consequently the expectations,
\begin{equation}
    \begin{split}
        &q(c_n^j=c|\z_n^j;\theta^j_{\text{old}})\\
        &=\frac{q(\z_n^j|c_n^j=c;\theta^j_{\text{old}})\cdot q(c_n^j=c;\theta^j_{\text{old}})}{q(\z_n^j;\theta^j_{\text{old}})}\\
        &=\frac{q(\z_n^j|c_n^j=c;\theta^j_{\text{old}})\cdot q(c_n^j=c;\theta^j_{\text{old}})}{\sum_{c=1}^C q(\z_n^j|c_n^j=c;\theta^j_{\text{old}})\cdot q(c_n^j=c;\theta^j_{\text{old}})}\\
        &=\frac{\pi_c^j \cdot \mathcal{N}(\z_n^j; \boldsymbol{\mu}_c^j, \Sigma_c^j)}{\sum_{c=1}^C \pi_c^j \cdot \mathcal{N}(\z_n^j; \boldsymbol{\mu}_c^j, \Sigma_c^j)}.\\
    \end{split}
\end{equation}
In the M-step, we find $\theta^j_{\text{new}}$ that maximizes the surrogate function. Since the entropy term $H(C;\theta^j_{\text{old}})$ is not a function of $\theta_j$ (it is a function of $\theta^j_{\text{old}}$), we only need to optimize $Q(\theta^j; \theta^j_{\text{old}})$ by taking the derivative with respect to $\theta_j$,
\begin{equation}
    \begin{split}
        &\sum_{n=1}^{N_j}\frac{\partial Q(\theta^j;\theta^j_{\text{old}})}{\partial \pi_c^j}=0\\
        &\Rightarrow \pi_c^{j, \text{new}} = \frac{\sum_{n=1}^{N_j} q(c_n^j=c|\z_n^j;\theta_{\text{old}}^j)}{N_j}\\
        &\sum_{n=1}^{N_j}\frac{\partial Q(\theta^j;\theta^j_{\text{old}})}{\partial \boldsymbol{\mu}_c^j}=0\\
        &\Rightarrow \boldsymbol{\mu}_c^{j,\text{new}}=\frac{\sum_{n=1}^{N_j} q(c_n^j=c|\z_n^j;\theta_{\text{old}}^j)\cdot\z_n^j}{\sum_{n=1}^{N_j} q(c_n^j=c|\z_n^j;\theta_{\text{old}}^j)} \\
        &\sum_{n=1}^{N_j}\frac{\partial Q(\theta^j;\theta^j_{\text{old}})}{\partial \Sigma_c^j}=0\\
        &\Rightarrow \Sigma_c^{j,\text{new}}=\frac{\sum_{n=1}^{N_j} q(c_n^j=c|\z_n^j;\theta_{\text{old}}^j)\cdot (\z_n^j - \boldsymbol{\mu}_c^{j,\text{new}})^2}{\sum_{n=1}^{N_j} q(c_n^j=c|\z_n^j;\theta_{\text{old}}^j)}.\\
    \end{split}
\end{equation}

\section{Training details} For training, we use weight decay of $1\times 10^{-5}$ and AdamW optimizer with a learning rate of $1\times 10^{-4}$ with the cosine decay scheduler. For \textit{slide classification} experiments, we use cross-entropy loss and a maximum of 20 epochs with early stopping if the validation loss does not decrease for 10 epochs. For the supervised baselines, due to the variable-length WSI set, we use a batch size of 1 and a gradient accumulation of 32 steps. For unsupervised baselines (including $\ours$), we use a batch size of 32. For \textit{survival prediction} experiments, we use negative log-likelihood loss (NLL)~\cite{zadeh2020bias} with a batch size of one patient over 20 epochs for supervised baselines. For unsupervised baselines (including $\ours$), we use Cox proportional hazards loss~\cite{katzman2018deepsurv} with a batch size of 64 patients over 50 epochs.

\section{Computational considerations}
Two NVIDIA 3090 GPUs were used for training $\ours$. $\ours$ pre-extracts 32,784-dim slide features (16 prototypes $\times$ 2,049-dim for concatenated $\pi_c$, $\boldsymbol{\mu}_c, \Sigma_c$) for linear or MLP probing, \textbf{468$\times$} smaller than $[15\text{K} \times 1024]$-dim patch patch embeddings used for MIL training. We pre-extract $\ours$ features with batch size of 1 (10 WSIs/sec), and can compress 11K TCGA slides (4 TB) with $\sim$ 1.4 GB. While more prototypes imply more features to concatenate, training a linear classifier with $\ours$ features still has less number of parameters (32,784) than ABMIL ($\sim$500K). 

\section{Datasets}
We provide brief explanations for the datasets that were used for the evaluation of $\ours$.

\subsection{Slide classification}
\textbf{EBRAINS}~\cite{roetzer2022digital}: For fine-grained (30 classes) and coarse-grained (12 classes) brain tumor subtyping tasks, we used Hematoxylin and Eosin (H\&E) Formalin-fixed and paraffin-embedded (FFPE) WSIs ($n=2,319$) collected from the University of Vienna. We label-stratify the dataset into train/val/test fold of 50:25:25 and use the same fold for both the fine-grained and coarse-grained subtyping tasks. Performance was evaluated using balanced accuracy and weighted F1.

\noindent\textbf{NSCLC}: For the non-small cell lung carcinoma (NSCLC) subtyping task, we use H\&E WSIs from TCGA and CPTAC for classifying lung adenocarcinoma (LUAD) and lung squamous cell carcinoma (LUSC) cases. The TCGA cohort contains a total of 1,041 slides (LUAD: 529, LUSC: 512) and the CPTAC cohort consists a total of 1,091 slides (LUAD: 578, LUSC: 513). We label-stratify the TCGA cohort into train/val/test fold of 80:10:10, with CPTAC used for external validation. Performance was evaluated using balanced accuracy and weighted F1.

\noindent\textbf{PANDA}~\cite{bulten2022artificial, bulten2020automated}: For the ISUP grading task, we used prostate cancer core needle biopsies (n=10,616) from the Prostate Cancer Grade Assessment (PANDA) challenge. Each biopsy is given an ISUP grade, making this a 6-class classification task. These biopsies are collected from Karolinska Institute (KRLS) and Radboud University Medical Center (RUMC). 
We label-stratify the PANDA dataset into train/val/test fold of 80:10:10,
with the evaluation performed on KRLS and RUMC cohorts separately. Performance was evaluated using Cohen's quadratic weighted Kappa $\kappa^2$ metric.

\subsection{Survival prediction}
\textbf{TCGA}: We perform site-stratified 5-fold CV~\cite{howard2021impact} evaluation on the following cancer types from TCGA: Breast
Invasive Carcinoma (BRCA, $n=1,041, \text{WSI}=1,111$), Colon and Rectum Adenocarcinoma (CRC, $n=566, \text{WSI}=575$), Bladder Urothelial Carcinoma (BLCA, $n=373, \text{WSI}=437$), Uterine corpus endometrial carcinoma (UCEC, $n=504, \text{WSI}=565$), Kidney renal clear cell carcinoma (KIRC, $n=511, \text{WSI}=517$), and Lung adenocarcinoma (LUAD, $n=456, \text{WSI}=1,024$). The train/val split is performed on the patient level.\\
\textbf{External dataset (CPTAC, NLST)}: Using the models trained on TCGA cohort, we perform external validation on KIRC (CPTAC: $n=180, \text{WSI}=341$) and LUAD (CPTAC: $n=185, \text{WSI}=486$, NLST: $n=244, \text{WSI}=686$). We note that evaluation on CPTAC and NLST is much more difficult due to dataset shifts in image acquisition (differences in H\&E stain variability), geographic location and demographics (social determinants of health affecting access to healthcare), and other potential biases (differences in follow-up procedures between TCGA and CPTAC/NLST).

\section{Additional experiments}

\noindent\textbf{Ablation over different feature encoders:} We evaluate $\ours$, which relied on features extracted with ViT-L/16 DINOv2 pre-trained on a large internal histology dataset (UNI)~\cite{chen2024towards}, with other baselines using features extracted from 1) CTransPath encoder~\cite{wang2022transformer}, which is a Swin Transformer pretrained on 29,753 WSIs from TCGA and 2,457 WSIs from the Pathology AI Platform (PAIP), and 2) ResNet50 encoder pretrained on natural images (ImageNet)~\cite{lu2021data}. The results can be found in Table~\ref{tab:ablation_extractor}.

\noindent\textbf{Ablation over a different number of clusters $C$:} We evaluate how $\ours$ and other baselines (AttnMISL, ProtoCounts, H2T, and OT) that depend on the number of prototypes $C$ perform across different choices. We report both the classification and survival prediction results for $C=\{8, 16, 32\}$ in Table~\ref{tab:ablation_c}. 

\noindent\textbf{Ablation over different survival loss functions:} For survival prediction tasks, we also train our unsupervised baselines with 1) the negative log-likelihood (NLL) loss~\cite{zadeh2020bias} and 2) the ranking loss~\cite{luck2018learning}. These survival loss functions have been frequently used as alternatives to the Cox loss in survival analysis problems, especially in medical imaging literature. To maintain consistency with the Cox loss experimental setting, we use a batch size of 64 for training with the NLL and ranking loss. The results can be found in Table~\ref{tab:ablation_loss}, where we include the supervised baseline results with NLL loss for completeness.

Evaluation was performed on several representative classification and survival tasks: EBRAINS (challenging difficulty), PANDA (depends on understanding mixture proportions of tissue patterns), CRC survival prediction (tissue can be annotated using CRC-100K~\cite{kather2019predicting}), and LUAD survival prediction (assessing out-of-domain generalization).

\section{Results, interpretation, and insights}

\noindent\textbf{Stronger feature encoders improve supervised MIL baselines:} We observe consistent trends that stronger feature encoders improve slide-level tasks, with models trained using UNI reaching the best performance (Table~\ref{tab:ablation_extractor}). Across all MIL architectures and in all classification tasks (except for DSMIL on RUMC evaluation in PANDA), UNI consistently outperforms ResNet-50 and CTransPath in head-to-head comparisons. On survival tasks, we note that CTransPath was additionally pretrained on TCGA, which may produce optimistic bias in evaluation. However, we find that UNI still outperforms CTransPath on TCGA-CRC and TCGA-LUAD survival prediction across many architectures. Interestingly, AttnMISL, which generally underperformed against ABMIL and TransMIL using ResNet-50 features, becomes one of the top-ranked MIL models in survival tasks when using UNI features (second-highest c-index in CRC survival prediction, and the highest c-index in LUAD survival prediction on the TCGA and NLST cohorts). This can be attributed to the prototypical formulation of AttnMISL, which depends on the representation quality of the data centroids for prototypical pooling, which would improve with stronger feature encoders.\\

\noindent\textbf{Stronger feature encoders enable unsupervised baselines to compete with MIL:} Similar to MIL methods, $\ours$ and other unsupervised slide representation methods have consistent improvement using UNI over ResNet-50 and CTransPath (aside from evaluation on PANDA and evaluation using DeepSets/ProtoCounts, Table~\ref{tab:ablation_extractor}). Interestingly, we note that OT and $\ours$ using CTransPath features significantly underperforms against many weakly-supervised baselines (0.377 / 0.369 / 0.518 balanced accuracies comparing OT, $\ours$, ABMIL respectively on EBRAINS; 0.677 / 0.782 / 0.901 $\kappa^2$ comparing OT, $\ours$, ABMIL respectively on PANDA-KRLS). When using a stronger feature encoder such as UNI, unsupervised slide representation methods have significant gains and consequently outperform ABMIL and other MIL baselines. As in the case of AttnMISL, this can be attributed to the need for strong pretrained encoders that are able to retrieve similar patch embeddings for prototypical pooling. \\


\noindent\textbf{$\ours$ trains stable survival models:} Across all evaluation settings (different number of prototypes $C$ and survival loss functions), we find that $\ours$ is able to develop high-performing survival models with out-of-domain generalization (Table~\ref{tab:ablation_c} and \ref{tab:ablation_loss}). In CRC survival prediction, $\ours_{\text{All}}$+MLP consistently outperforms all cluster-based methods within each setting. In comparison to MIL, though $\ours_{\text{All}}$+MLP with $C$=16 has lower c-index than TransMIL (0.684), we note that $\ours_{\text{All}}$+MLP with $C$=8 reaches higher performance (0.691). In LUAD survival prediction, $\ours_{\text{All}}$+MLP is consistently the second best-performing model on TCGA evaluation, behind OT (best c-index of 0.715 of with $C$=8). Though many baselines such as OT, H2T, and even DeepSets can reach strong performance on TCGA, we note that almost all of these methods have unstable performance on external cohorts, with c-index falling under 0.5 on CPTAC, NLST, or both.\\ 

\begin{figure*}[t]
   \centering
   \includegraphics[width=1\linewidth]{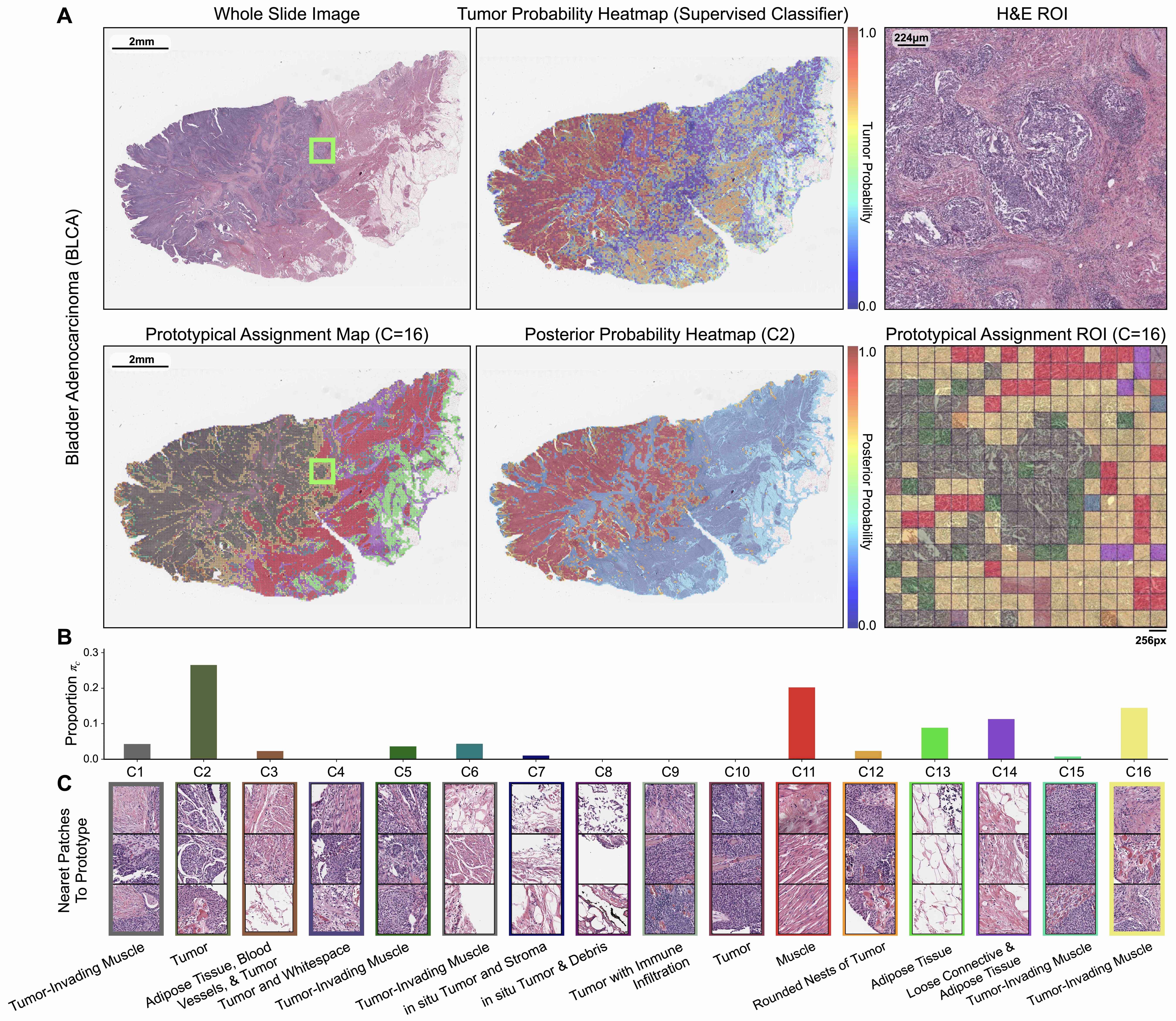}
   \caption{\textbf{Prototype-oriented heatmap interpretation of BLCA}. (A) Visualization of prototypical assignment map in an exemplar BLCA H\&E WSI, with zoomed-in histopathology ROI of tumor-invading muscle (C2, C8, C11, C16). We show the posterior probability heatmap for the tumor-containing C2 prototype, which has strong concordance with a tumor probability heatmap obtained by a supervised patch-level classifier for BLCA tumor prediction. (B) Prototype distribution $\hat{\pi}_c$ of the exemplar slide. (C) Morphological annotations of all prototypes by a board-certified pathologist in the BLCA cohort.
   }
   \label{fig:heatmap_blca}
\end{figure*}

\begin{figure*}[t]
   \centering
   \includegraphics[width=1\linewidth]{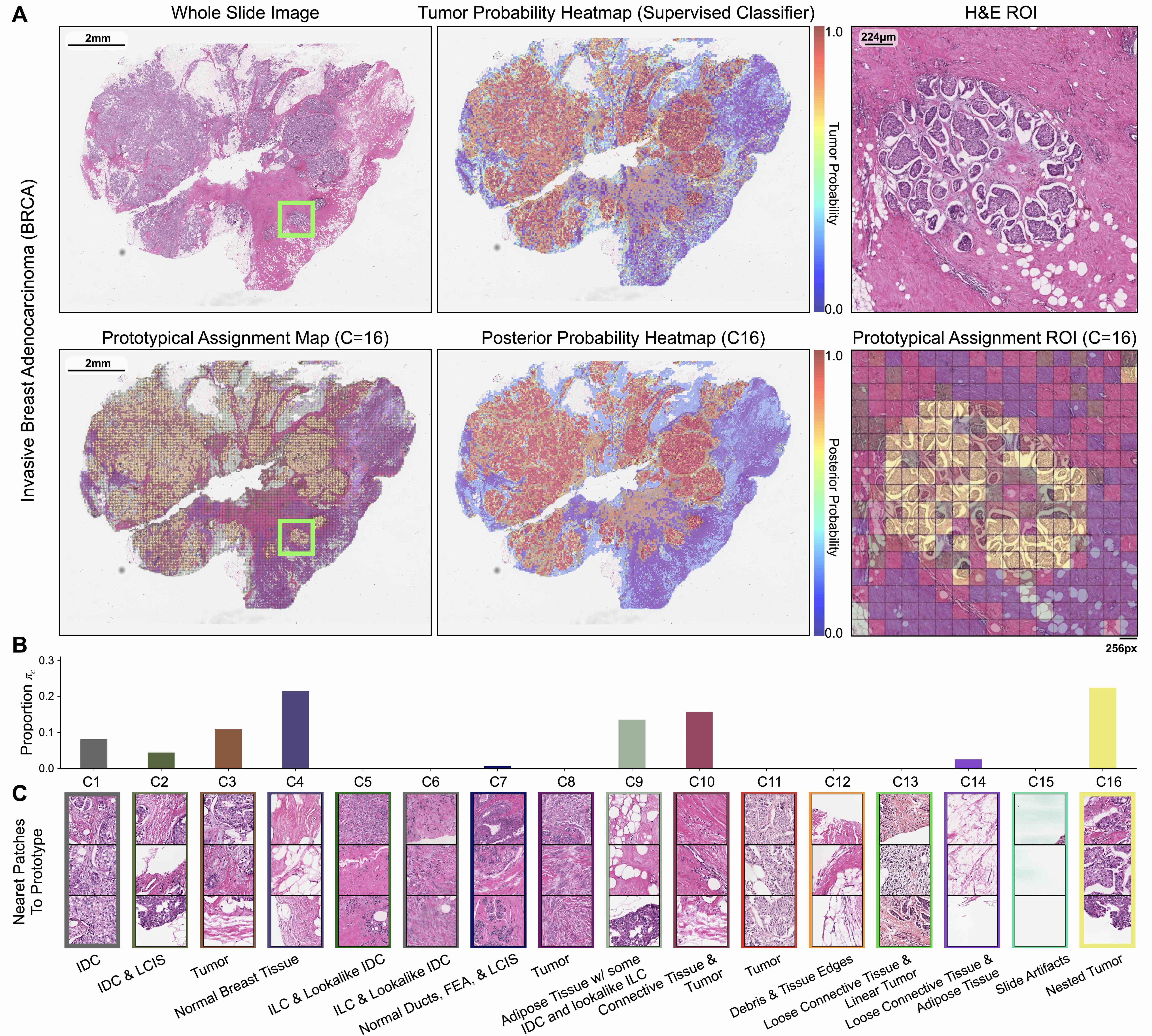}
   \caption{\textbf{Prototype-oriented heatmap interpretation of BRCA}. (A) Visualization of prototypical assignment map in an exemplar BRCA H\&E WSI, with zoomed-in histopathology ROI of dense tumor nests (C16) with surrounding connective tissue (C10), adipose tissue (C9) with tumor presence (C3). We show the posterior probability heatmap for the tumor-containing C16 prototype, which has strong concordance with a tumor probability heatmap obtained by a supervised patch-level classifier for BRCA tumor prediction. (B) Prototype distribution $\hat{\pi}_c$ of the exemplar slide. (C) Morphological annotations of all prototypes by a board-certified pathologist in the BRCA cohort.
   }
   \label{fig:heatmap_brca}
\end{figure*}

\begin{figure*}[t]
   \centering
   \includegraphics[width=1\linewidth]{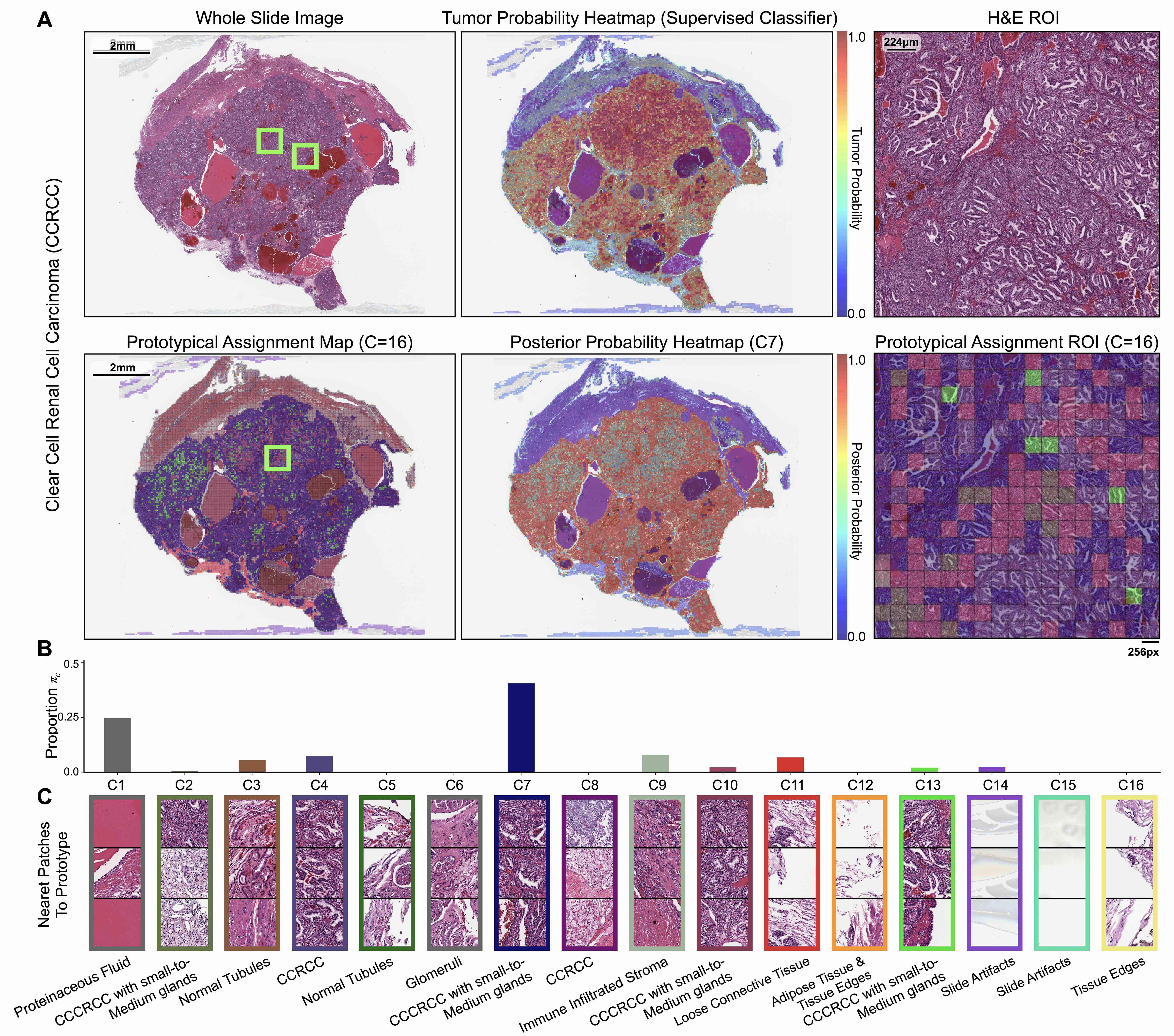}
   \caption{\textbf{Prototype-oriented heatmap interpretation of KIRC}. (A) Visualization of prototypical assignment map in an exemplar KIRC/CCRCC H\&E WSI, with zoomed-in histopathology ROI of CCRCC in small-to-medium glands (C7, C10, C13). We show the posterior probability heatmap for the tumor-containing C7 prototype, which has strong concordance with a tumor probability heatmap obtained by a supervised patch-level classifier for CCRCC tumor prediction. (B) Prototype distribution $\hat{\pi}_c$ of the exemplar slide. (C) Morphological annotations of all prototypes by a board-certified pathologist in the KIRC cohort.
   }
   \label{fig:heatmap_kirc}
\end{figure*}

\noindent\textbf{$\ours$ prototypes capture distinct tumor morphologies:} In \textbf{Fig.}~\ref{fig:heatmap_blca},~\ref{fig:heatmap_brca}, and ~\ref{fig:heatmap_kirc}, we visualize prototypical assignment maps and heatmap visualizations of various cancer types. Consistent with our findings in \textbf{Fig.}~\ref{fig:heatmap}, $\ours$ is able to map the spatial organization of histologic visual concepts. In particular, $\ours$ finds several unique tumor populations, delineating: tumor-invading muscle and tumor with immune infiltration in BLCA, nested tumor and tumor-associated connective tissue in BRCA, and clear cell RCC with and without presentation of poorly-differentiated glands in KIRC. Furthermore, we also show concordance of our visualizations using a supervised classifier (developed using patch-level tumor annotations in the TCGA Uniform Tumor dataset~\cite{komura2022universal}) for tumor tissue classification. Visualizing the posterior probability heatmap of the tumor prototype with the highest mixture probability $\widehat{\pi}_c$ (greatest presence), we find that our tumor heatmap visualizations have strong concordance with those generated based on the results from supervised classifiers.\\

\begin{figure*}[t]
   \centering
   \includegraphics[width=1\linewidth]{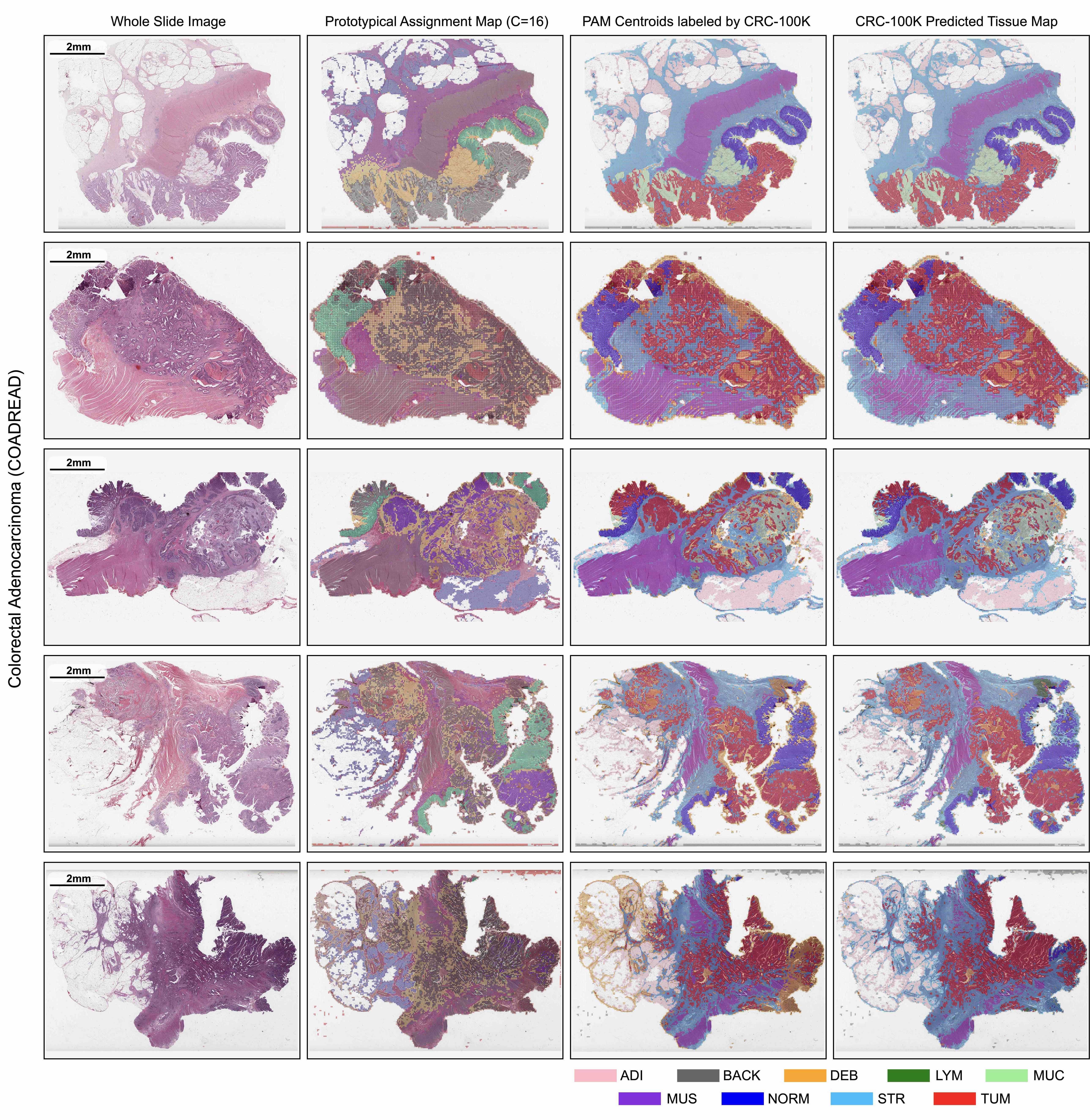}
   \caption{\textbf{Prototype-oriented heatmap interpretation of COADREAD and correspondence with CRC-100K}. For exemplar COADREAD slides, we visualize their prototypical assignment maps and their correspondence with tissue classes in CRC-100K. Using a supervised patch-level classifier for predicting the 9 tissue classes in CRC-100K, we predicted tissue classes in TCGA-COADREAD slides, shown in the \textbf{far-right column}. To match the prototypical assignment maps from $\ours$ with the label distribution in CRC-100K, we applied the same classifier to predict CRC-100K tissue labels for the learned prototypes in $\ours$ (\textbf{middle-right column}). Across all 9 classes, we find that $\ours$'s prototypes correspond to morphologically-relevant and semantic histopathology tissue patterns annotated by supervised classifiers.
   }
   \label{fig:heatmap_crc}
\end{figure*}

\noindent\textbf{$\ours$ prototypes capture distribution of tissue classes in CRC-100K:} In \textbf{Fig.}~\ref{fig:heatmap_crc}, we visualize prototypical assignment maps and their correspondence to diverse tissue annotations in COADREAD tissue. Using the CRC-100K dataset (containing 9 tissue classes)~\cite{kather2019predicting}, we developed a supervised patch-level classifier to predict tissue assignments for all patches in TCGA-COADREAD slides. To match the prototypical assignment maps from $\ours$ with the label distribution in CRC-100K, we applied the previous classifier to the learned prototypes of $\ours$, to predict CRC-100K tissue labels. Overall, we find that the learned prototypes of $\ours$ have strong concordance with morphologically-relevant and diverse histopathology tissue patterns annotated by supervised classifiers.

\clearpage

\clearpage

\begin{table*}[!ht]
\centering
\small
\caption{\textbf{Varying pretrained feature extractors.} We compare the performance of supervised (\textbf{top}) and unsupervised (\textbf{bottom}) methods with different pretrained encoders, ResNet50 with ImageNet transfer (RN50), CTransPath (CTP), and UNI, on classification tasks (EBRAINS and PANDA) and survival tasks (CRC and LUAD).}
\begin{tabular}{ll|cc|cc|c|ccc}
\toprule
&\textbf{Train on} & \multicolumn{2}{c}{EBRAINS} & \multicolumn{2}{c}{PANDA} & \multicolumn{1}{c}{CRC} & \multicolumn{3}{c}{LUAD} \\
& & \multicolumn{2}{c}{(fine, 30 classes)} & \multicolumn{2}{c}{(grading, 6 classes)} & \multicolumn{1}{c}{(survival)} & \multicolumn{3}{c}{(survival)} \\
\cline{3-10}
&\textbf{Test on} & \multicolumn{2}{c|}{EBRAINS} & KRLS & RUMC & TCGA & TCGA & CPTAC & NLST \\
&& (Bal. acc.) & (F1) & ($\kappa^2$) & ($\kappa^2$) & (C-Index) & (C-Index) & (C-Index) & (C-Index) \\
\midrule
\parbox[t]{0mm}{\multirow{5}{*}{\rotatebox[origin=c]{90}{{ \textbf{Sup. (RN50)} }}}} 
& ABMIL~\cite{ilse2018attention}& 0.197 & 0.267 & 0.792 & 0.789 & 0.540 $\pm$ 0.08 & 0.537 $\pm$ 0.18 & 0.684 $\pm$ 0.01 & 0.482 $\pm$ 0.03 \\
& TransMIL~\cite{shao2021transmil}& 0.516 & 0.614 & 0.841 & 0.854 & 0.501 $\pm$ 0.05 & 0.628 $\pm$ 0.11 & 0.555 $\pm$ 0.06 & 0.462 $\pm$ 0.01 \\
& DSMIL~\cite{li2021dual}& 0.401 & 0.510 & 0.681 & 0.728 & 0.494 $\pm$ 0.01 & 0.487 $\pm$ 0.03 & 0.524 $\pm$ 0.02 & 0.519 $\pm$ 0.02 \\
& ILRA~\cite{xiang2023exploring}& 0.509 & 0.599 & 0.860 & 0.880 & 0.588 $\pm$ 0.09 & 0.522 $\pm$ 0.17 & 0.559 $\pm$ 0.07 & 0.443 $\pm$ 0.06 \\
&AttnMISL~\cite{yao2020whole}& 0.033 & 0.073 & 0.128 & 0.005 & 0.493 $\pm$ 0.04 & 0.519 $\pm$ 0.16 & 0.666 $\pm$ 0.01 & 0.535 $\pm$ 0.01 \\
\midrule
\parbox[t]{0mm}{\multirow{5}{*}{\rotatebox[origin=c]{90}{{ \textbf{Sup. (CTP)} }}}} 
& ABMIL~\cite{ilse2018attention}& 0.518 & 0.594 & 0.901 & 0.908 & 0.642 $\pm$ 0.10 & 0.607 $\pm$ 0.03 & 0.545 $\pm$ 0.05 & 0.560 $\pm$ 0.02 \\
& TransMIL~\cite{shao2021transmil}& 0.642 & 0.71 & 0.911 & 0.918 & 0.611 $\pm$ 0.12 & 0.614 $\pm$ 0.06 & 0.479 $\pm$ 0.05 & 0.508 $\pm$ 0.04 \\
& DSMIL~\cite{li2021dual}& 0.515 & 0.584 & 0.890 & 0.916 & 0.499 $\pm$ 0.03 & 0.552 $\pm$ 0.05 & 0.466 $\pm$ 0.04 & 0.438 $\pm$ 0.01 \\
& ILRA~\cite{xiang2023exploring}& 0.580 & 0.655 & 0.917 & 0.920 & 0.590 $\pm$ 0.09 & 0.602 $\pm$ 0.05 & 0.427 $\pm$ 0.05 & 0.456 $\pm$ 0.03 \\
&AttnMISL~\cite{yao2020whole}& 0.033 & 0.073 & 0.402 & 0.837 & 0.627 $\pm$ 0.12 & 0.602 $\pm$ 0.05 & 0.427 $\pm$ 0.05 & 0.456 $\pm$ 0.03 \\
\midrule
\parbox[t]{0mm}{\multirow{5}{*}{\rotatebox[origin=c]{90}{{ \textbf{Sup. (UNI)} }}}} 
& ABMIL~\cite{ilse2018attention}& 0.674 & 0.744 & 0.935 & 0.918 & 0.608 $\pm$ 0.09 & 0.654 $\pm$ 0.06 & 0.572 $\pm$ 0.03 & 0.519 $\pm$ 0.04 \\
& TransMIL~\cite{shao2021transmil}& 0.701 & 0.758 & 0.942 & 0.922 & 0.684 $\pm$ 0.06 & 0.665 $\pm$ 0.10 & 0.555 $\pm$ 0.03 & 0.484 $\pm$ 0.05 \\
& DSMIL~\cite{li2021dual}& 0.648 & 0.698 & 0.909 & 0.911 & 0.500 $\pm$ 0.00 & 0.501 $\pm$ 0.00 & 0.502 $\pm$ 0.00 & 0.500 $\pm$ 0.00 \\
& ILRA~\cite{xiang2023exploring}& 0.618 & 0.695 & 0.931 & 0.925 & 0.555 $\pm$ 0.10 & 0.586 $\pm$ 0.06 & 0.651 $\pm$ 0.05 & 0.482 $\pm$ 0.01 \\
&AttnMISL~\cite{yao2020whole}& 0.534 & 0.636 & 0.882 & 0.894 & 0.639 $\pm$ 0.10 & 0.673 $\pm$ 0.10 & 0.632 $\pm$ 0.03 & 0.577 $\pm$ 0.04 \\
\midrule\midrule
\parbox[t]{0mm}{\multirow{7}{*}{\rotatebox[origin=c]{90}{{ \textbf{Unsup. (RN50)} }}}}
& DeepSets~\cite{zaheer2017deep} & 0.033 & 0.073 & $<0$ & $<0$ & 0.574 $\pm$ 0.08 & 0.565 $\pm$ 0.12 & 0.715 $\pm$ 0.02 & 0.591 $\pm$ 0.01 \\
&ProtoCounts~\cite{yu2023prototypical, quiros2023mapping}& 0.045 & 0.077 & 0.016 & 0.183 & 0.516 $\pm$ 0.05 & 0.546 $\pm$ 0.04 & 0.499 $\pm$ 0.08 & 0.502 $\pm$ 0.07 \\
&H2T~\cite{VU2023handcrafted}& 0.047 & 0.087 & 0.262 & 0.329 & 0.501 $\pm$ 0.10 & 0.585 $\pm$ 0.14 & 0.512 $\pm$ 0.07 & 0.545 $\pm$ 0.04 \\
&OT~\cite{mialon2021a}& 0.063 & 0.088 & 0.211 & 0.540 & 0.578 $\pm$ 0.08 & 0.575 $\pm$ 0.14 & 0.581 $\pm$ 0.05 & 0.544 $\pm$ 0.02 \\
&$\ours_{\text{WA}}$ + lin.& 0.033 & 0.073 & 0.150 & 0.057 & 0.525 $\pm$ 0.09 & 0.586 $\pm$ 0.11 & 0.552 $\pm$ 0.04 & 0.491 $\pm$ 0.01 \\
&$\ours_{\text{All}}$ + lin.& 0.063 & 0.112 & 0.207 & 0.535 & 0.554 $\pm$ 0.07 & 0.586 $\pm$ 0.12 & 0.548 $\pm$ 0.04 & 0.505 $\pm$ 0.02 \\
&$\ours_{\text{All}}$ + MLP& 0.142 & 0.216 & 0.550 & 0.665 & 0.585 $\pm$ 0.07 & 0.601 $\pm$ 0.07 & 0.396 $\pm$ 0.04 & 0.465 $\pm$ 0.03 \\
\midrule
\parbox[t]{0mm}{\multirow{7}{*}{\rotatebox[origin=c]{90}{{ \textbf{Unsup. (CTP)} }}}}
& DeepSets~\cite{zaheer2017deep} & 0.033 & 0.073 & $<0$ & $<0$ & 0.522 $\pm$ 0.10 & 0.629 $\pm$ 0.06 & 0.463 $\pm$ 0.03 & 0.554 $\pm$ 0.05 \\
&ProtoCounts~\cite{yu2023prototypical, quiros2023mapping}& 0.057 & 0.037 & 0.059 & 0.539 & 0.521 $\pm$ 0.05 & 0.479 $\pm$ 0.11 & 0.537 $\pm$ 0.10 & 0.522 $\pm$ 0.11 \\
&H2T~\cite{VU2023handcrafted}& 0.053 & 0.124 & 0.333 & 0.704 & 0.545 $\pm$ 0.11 & 0.586 $\pm$ 0.05 & 0.552 $\pm$ 0.07 & 0.562 $\pm$ 0.02 \\
&OT~\cite{mialon2021a}& 0.377 & 0.482 & 0.677 & 0.738 & 0.607 $\pm$ 0.09 & 0.690 $\pm$ 0.06 & 0.466 $\pm$ 0.02 & 0.547 $\pm$ 0.06 \\
&$\ours_{\text{WA}}$ + lin.& 0.033 & 0.073 & 0.203 & 0.586 & 0.533 $\pm$ 0.11 & 0.673 $\pm$ 0.05 & 0.474 $\pm$ 0.03 & 0.515 $\pm$ 0.05 \\
&$\ours_{\text{All}}$ + lin.& 0.398 & 0.493 & 0.661 & 0.757 & 0.614 $\pm$ 0.09 & 0.672 $\pm$ 0.05 & 0.491 $\pm$ 0.04 & 0.540 $\pm$ 0.06 \\
&$\ours_{\text{All}}$ + MLP& 0.369 & 0.483 & 0.782 & 0.869 & 0.661 $\pm$ 0.11 & 0.655 $\pm$ 0.08 & 0.584 $\pm$ 0.04 & 0.511 $\pm$ 0.03 \\
\midrule
\parbox[t]{0mm}{\multirow{7}{*}{\rotatebox[origin=c]{90}{{ \textbf{Unsup. (UNI)} }}}}
& DeepSets~\cite{zaheer2017deep} & 0.033 & 0.073 & $<$ 0 & $<$ 0 & 0.563 $\pm$ 0.10 & 0.652 $\pm$ 0.05 & 0.550 $\pm$ 0.01 & 0.509 $\pm$ 0.04 \\
&ProtoCounts~\cite{yu2023prototypical, quiros2023mapping}& 0.038 & 0.018 & $<0$ & 0.13 & 0.552 $\pm$ 0.06 & 0.460 $\pm$ 0.11 & 0.577 $\pm$ 0.11 & 0.500 $\pm$ 0.01 \\
&H2T~\cite{VU2023handcrafted}& 0.117 & 0.223 & 0.457 & 0.755 & 0.639 $\pm$ 0.11 & 0.662 $\pm$ 0.09 & 0.583 $\pm$ 0.03 & 0.603 $\pm$ 0.04 \\
&OT~\cite{mialon2021a}& 0.700 & 0.756 & 0.817 & 0.883 & 0.622 $\pm$ 0.09 & 0.687 $\pm$ 0.08 & 0.641 $\pm$ 0.02 & 0.495 $\pm$ 0.04 \\
&$\ours_{\text{WA}}$ + lin.& 0.497 & 0.598 & 0.663 & 0.787 & 0.647 $\pm$ 0.12 & 0.654 $\pm$ 0.07 & 0.461 $\pm$ 0.01 & 0.482 $\pm$ 0.06 \\
&$\ours_{\text{All}}$ + lin.& 0.691 & 0.756 & 0.866 & 0.909 & 0.645 $\pm$ 0.07 & 0.672 $\pm$ 0.06 & 0.568 $\pm$ 0.05 & 0.623 $\pm$ 0.07 \\
&$\ours_{\text{All}}$ + MLP& 0.693 & 0.760 & 0.923 & 0.931 & 0.665 $\pm$ 0.10 & 0.685 $\pm$ 0.06 & 0.653 $\pm$ 0.04 & 0.634 $\pm$ 0.04 \\
\bottomrule
\end{tabular}
\label{tab:ablation_extractor}
\end{table*}

\clearpage

\begin{table*}[!ht]
\centering
\small
\caption{\textbf{Varying $C$ in cluster-based methods.} We compare the performance of cluster-based methods with $C=\{8,16,32\}$ on classification tasks (EBRAINS and PANDA) and survival tasks (CRC and LUAD). \textbf{Top.} MIL baselines with no clustering (NC). \textbf{Bottom.} Cluster-based methods with $C=\{8,16,32\}$, which include weakly-supervised MIL (AttnMISL) and unsupervised slide representation learning approaches (ProtoCounts, H2T, OT, PANTHER).}
\begin{tabular}{ll|cc|cc|c|ccc}
\toprule
&\textbf{Train on} & \multicolumn{2}{c}{EBRAINS} & \multicolumn{2}{c}{PANDA} & \multicolumn{1}{c}{CRC} & \multicolumn{3}{c}{LUAD} \\
& & \multicolumn{2}{c}{(fine, 30 classes)} & \multicolumn{2}{c}{(grading, 6 classes)} & \multicolumn{1}{c}{(survival)} & \multicolumn{3}{c}{(survival)} \\
\cline{3-10}
&\textbf{Test on} & \multicolumn{2}{c|}{EBRAINS} & KRLS & RUMC & TCGA & TCGA & CPTAC & NLST \\
&& (Bal. acc.) & (F1) & ($\kappa^2$) & ($\kappa^2$) & (C-Index) & (C-Index) & (C-Index) & (C-Index) \\
\midrule
\parbox[t]{0mm}{\multirow{5}{*}{\rotatebox[origin=c]{90}{{\textbf{NC}}}}}
& ABMIL~\cite{ilse2018attention}& 0.674 & 0.744 & 0.935 & 0.918 & 0.608 $\pm$ 0.09 & 0.654 $\pm$ 0.06 & 0.572 $\pm$ 0.03 & 0.519 $\pm$ 0.04 \\
& TransMIL~\cite{shao2021transmil}& 0.701 & 0.758 & 0.942 & 0.922 & 0.684 $\pm$ 0.06 & 0.665 $\pm$ 0.10 & 0.555 $\pm$ 0.03 & 0.484 $\pm$ 0.05 \\
& DSMIL~\cite{li2021dual}& 0.648 & 0.698 & 0.909 & 0.911 & 0.500 $\pm$ 0.00 & 0.501 $\pm$ 0.00 & 0.502 $\pm$ 0.00 & 0.50 $\pm$ 0.00 \\
& ILRA~\cite{xiang2023exploring}& 0.618 & 0.695 & 0.931 & 0.925 & 0.555 $\pm$ 0.10 & 0.586 $\pm$ 0.06 & 0.651 $\pm$ 0.05 & 0.482 $\pm$ 0.01 \\
& DeepSets~\cite{zaheer2017deep} & 0.033 & 0.073 & $<$ 0 & $<$ 0 & 0.563 $\pm$ 0.10 & 0.652 $\pm$ 0.05 & 0.550 $\pm$ 0.01 & 0.509 $\pm$ 0.04 \\
\midrule
\parbox[t]{0mm}{\multirow{7}{*}{\rotatebox[origin=c]{90}{{\textbf{C=8}}}}}
&AttnMISL~\cite{yao2020whole}& 0.560 & 0.655 & 0.857 & 0.874 & 0.595 $\pm$ 0.08 & 0.643 $\pm$ 0.07 & 0.644 $\pm$ 0.04 & 0.545 $\pm$ 0.01 \\
&ProtoCounts~\cite{yu2023prototypical, quiros2023mapping}& 0.022 & 0.03 & 0.0 & 0.284 & 0.479 $\pm$ 0.11 & 0.561 $\pm$ 0.06 & 0.625 $\pm$ 0.13 & 0.562 $\pm$ 0.15 \\
&H2T~\cite{VU2023handcrafted}& 0.045 & 0.105 & 0.286 & 0.767 & 0.622 $\pm$ 0.09 & 0.638 $\pm$ 0.08 & 0.525 $\pm$ 0.04 & 0.563 $\pm$ 0.03 \\
&OT~\cite{mialon2021a}& 0.689 & 0.756 & 0.803 & 0.869 & 0.626 $\pm$ 0.09 & 0.715 $\pm$ 0.10 & 0.609 $\pm$ 0.03 & 0.523 $\pm$ 0.05 \\
&$\ours_{\text{WA}}$ + lin.& 0.490 & 0.593 & 0.671 & 0.787 & 0.600 $\pm$ 0.10 & 0.670 $\pm$ 0.05 & 0.526 $\pm$ 0.01 & 0.502 $\pm$ 0.07 \\
&$\ours_{\text{All}}$ + lin.& 0.668 & 0.742 & 0.843 & 0.9 & 0.678 $\pm$ 0.12 & 0.643 $\pm$ 0.06 & 0.660 $\pm$ 0.02 & 0.615 $\pm$ 0.03 \\
&$\ours_{\text{All}}$ + MLP& 0.674 & 0.753 & 0.918 & 0.936 & 0.691 $\pm$ 0.11 & 0.648 $\pm$ 0.06 & 0.669 $\pm$ 0.03 & 0.603 $\pm$ 0.03 \\
\midrule
\parbox[t]{0mm}{\multirow{7}{*}{\rotatebox[origin=c]{90}{{\textbf{C=16}}}}}
&AttnMISL~\cite{yao2020whole}& 0.534 & 0.636 & 0.882 & 0.894 & 0.639 $\pm$ 0.10 & 0.673 $\pm$ 0.10 & 0.632 $\pm$ 0.03 & 0.577 $\pm$ 0.04 \\
&ProtoCounts~\cite{yu2023prototypical, quiros2023mapping}& 0.038 & 0.018 & $<0$ & 0.13 & 0.552 $\pm$ 0.06 & 0.460 $\pm$ 0.11 & 0.577 $\pm$ 0.11 & 0.500 $\pm$ 0.01 \\
&H2T~\cite{VU2023handcrafted}& 0.117 & 0.223 & 0.457 & 0.755 & 0.639 $\pm$ 0.11 & 0.662 $\pm$ 0.09 & 0.583 $\pm$ 0.03 & 0.603 $\pm$ 0.04 \\
&OT~\cite{mialon2021a}& 0.700 & 0.756 & 0.817 & 0.883 & 0.622 $\pm$ 0.09 & 0.687 $\pm$ 0.08 & 0.641 $\pm$ 0.02 & 0.495 $\pm$ 0.04 \\
&$\ours_{\text{WA}}$ + lin.& 0.497 & 0.598 & 0.663 & 0.787 & 0.647 $\pm$ 0.12 & 0.654 $\pm$ 0.07 & 0.461 $\pm$ 0.01 & 0.482 $\pm$ 0.06 \\
&$\ours_{\text{All}}$ + lin.& 0.691 & 0.756 & 0.866 & 0.909 & 0.645 $\pm$ 0.07 & 0.672 $\pm$ 0.06 & 0.568 $\pm$ 0.05 & 0.623 $\pm$ 0.07 \\
&$\ours_{\text{All}}$ + MLP& 0.693 & 0.760 & 0.923 & 0.931 & 0.665 $\pm$ 0.10 & 0.685 $\pm$ 0.06 & 0.653 $\pm$ 0.04 & 0.634 $\pm$ 0.04 \\
\midrule
\parbox[t]{0mm}{\multirow{7}{*}{\rotatebox[origin=c]{90}{{\textbf{C=32}}}}}
&AttnMISL~\cite{yao2020whole}& 0.492 & 0.598 & 0.901 & 0.889 & 0.572 $\pm$ 0.10 & 0.666 $\pm$ 0.06 & 0.591 $\pm$ 0.03 & 0.587 $\pm$ 0.02 \\
&ProtoCounts~\cite{yu2023prototypical, quiros2023mapping}& 0.073 & 0.105 & 0.301 & 0.54 & 0.578 $\pm$ 0.09 & 0.498 $\pm$ 0.14 & 0.566 $\pm$ 0.12 & 0.529 $\pm$ 0.09 \\
&H2T~\cite{VU2023handcrafted}& 0.244 & 0.363 & 0.626 & 0.779 & 0.621 $\pm$ 0.12 & 0.665 $\pm$ 0.05 & 0.599 $\pm$ 0.04 & 0.650 $\pm$ 0.03 \\
&OT~\cite{mialon2021a}& 0.687 & 0.746 & 0.841 & 0.898 & 0.605 $\pm$ 0.00 & 0.689 $\pm$ 0.08 & 0.664 $\pm$ 0.02 & 0.518 $\pm$ 0.04 \\
&$\ours_{\text{WA}}$ + lin.& 0.489 & 0.593 & 0.670 & 0.782 & 0.606 $\pm$ 0.11 & 0.677 $\pm$ 0.06 & 0.522 $\pm$ 0.01 & 0.469 $\pm$ 0.06 \\
&$\ours_{\text{All}}$ + lin.& 0.676 & 0.751 & 0.883 & 0.896 & 0.649 $\pm$ 0.07 & 0.677 $\pm$ 0.06 & 0.583 $\pm$ 0.06 & 0.594 $\pm$ 0.05 \\
&$\ours_{\text{All}}$ + MLP& 0.674 & 0.741 & 0.935 & 0.931 & 0.656 $\pm$ 0.13 & 0.676 $\pm$ 0.04 & 0.665 $\pm$ 0.06 & 0.614 $\pm$ 0.05 \\
\bottomrule
\end{tabular}
\label{tab:ablation_c}
\end{table*}

\clearpage

\begin{table*}[!ht]
\centering
\small
\caption{\textbf{Varying loss function in survival tasks.} We compare the performance of all methods with different loss functions, NLL (\textbf{top}), ranking (\textbf{middle}), and Cox loss (\textbf{bottom}), on survival outcome prediction in CRC and LUAD.}
\begin{tabular}{ll|c|ccc}
\toprule
&\textbf{Train on} & \multicolumn{1}{c|}{CRC} & \multicolumn{3}{c}{LUAD} \\
&\textbf{Test on} & TCGA & TCGA & CPTAC & NLST \\
\midrule
\parbox[t]{0mm}{\multirow{5}{*}{\rotatebox[origin=c]{90}{{ \textbf{Sup. (NLL)} }}}} 
& ABMIL~\cite{ilse2018attention}& 0.608 $\pm$ 0.09 & 0.654 $\pm$ 0.06 & 0.572 $\pm$ 0.03 & 0.519 $\pm$ 0.04 \\
& TransMIL~\cite{shao2021transmil}& 0.684 $\pm$ 0.06 & 0.665 $\pm$ 0.10 & 0.555 $\pm$ 0.03 & 0.484 $\pm$ 0.05 \\
& DSMIL~\cite{li2021dual}& 0.500 $\pm$ 0.00 & 0.501 $\pm$ 0.00 & 0.502 $\pm$ 0.00 & 0.500 $\pm$ 0.00 \\
& ILRA~\cite{xiang2023exploring}& 0.555 $\pm$ 0.10 & 0.586 $\pm$ 0.06 & 0.651 $\pm$ 0.05 & 0.482 $\pm$ 0.01 \\
&AttnMISL~\cite{yao2020whole}& 0.639 $\pm$ 0.10 & 0.673 $\pm$ 0.10 & 0.632 $\pm$ 0.03 & 0.577 $\pm$ 0.04 \\
\midrule\midrule
\parbox[t]{0mm}{\multirow{7}{*}{\rotatebox[origin=c]{90}{{ \textbf{Unsup. (NLL)} }}}}
& DeepSets~\cite{zaheer2017deep} & 0.559 $\pm$ 0.11 & 0.560 $\pm$ 0.19 & 0.659 $\pm$ 0.02 & 0.587 $\pm$ 0.02 \\
&ProtoCounts~\cite{yu2023prototypical, quiros2023mapping}& 0.517 $\pm$ 0.03 & 0.493 $\pm$ 0.04 & 0.496 $\pm$ 0.15 & 0.591 $\pm$ 0.05 \\
&H2T~\cite{VU2023handcrafted}&  0.563 $\pm$ 0.08 & 0.498 $\pm$ 0.17 & 0.547 $\pm$ 0.01 & 0.520 $\pm$ 0.02 \\
&OT~\cite{mialon2021a} & 0.626 $\pm$ 0.12 & 0.681 $\pm$ 0.09 & 0.615 $\pm$ 0.03 & 0.462 $\pm$ 0.02 \\
&$\ours_{\text{WA}}$ + lin. &  0.508 $\pm$ 0.10 & 0.647 $\pm$ 0.09 & 0.643 $\pm$ 0.02 & 0.433 $\pm$ 0.04 \\
&$\ours_{\text{All}}$ + lin. & 0.647 $\pm$ 0.11 & 0.670 $\pm$ 0.08 & 0.651 $\pm$ 0.04 & 0.614 $\pm$ 0.07 \\
&$\ours_{\text{All}}$ + MLP &  0.649 $\pm$ 0.11 & 0.668 $\pm$ 0.08 & 0.638 $\pm$ 0.08 & 0.607 $\pm$ 0.05 \\
\midrule
\parbox[t]{0mm}{\multirow{7}{*}{\rotatebox[origin=c]{90}{{ \textbf{Unsup. (Rank)} }}}}
& DeepSets~\cite{zaheer2017deep} & 0.608 $\pm$ 0.11 & 0.614 $\pm$ 0.05 & 0.556 $\pm$ 0.04 & 0.538 $\pm$ 0.04     \\
&ProtoCounts~\cite{yu2023prototypical, quiros2023mapping} & 0.476 $\pm$ 0.05 & 0.503 $\pm$ 0.05 & 0.482 $\pm$ 0.14 & 0.507 $\pm$ 0.08 \\
&H2T~\cite{VU2023handcrafted} & 0.598 $\pm$ 0.11 & 0.661 $\pm$ 0.11 & 0.558 $\pm$ 0.02 & 0.620 $\pm$ 0.04 \\
&OT~\cite{mialon2021a} & 0.670 $\pm$ 0.11 & 0.643 $\pm$ 0.03 & 0.595 $\pm$ 0.03 & 0.488 $\pm$ 0.05 \\
&$\ours_{\text{WA}}$ + lin. & 0.626 $\pm$ 0.13 & 0.637 $\pm$ 0.06 & 0.445 $\pm$ 0.02 & 0.518 $\pm$ 0.07 \\
&$\ours_{\text{All}}$ + lin. & 0.661 $\pm$ 0.07 & 0.677 $\pm$ 0.06 & 0.575 $\pm$ 0.05 & 0.625 $\pm$ 0.06 \\
&$\ours_{\text{All}}$ + MLP & 0.671 $\pm$ 0.09 & 0.684 $\pm$ 0.06 & 0.651 $\pm$ 0.03 & 0.628 $\pm$ 0.05 \\
\midrule
\parbox[t]{0mm}{\multirow{7}{*}{\rotatebox[origin=c]{90}{{ \textbf{Unsup. (Cox)} }}}}
& DeepSets~\cite{zaheer2017deep} & 0.563 $\pm$ 0.10 & 0.652 $\pm$ 0.05 & 0.550 $\pm$ 0.01 & 0.509 $\pm$ 0.04 \\
&ProtoCounts~\cite{yu2023prototypical, quiros2023mapping}& 0.552 $\pm$ 0.06 & 0.460 $\pm$ 0.11 & 0.577 $\pm$ 0.11 & 0.500 $\pm$ 0.01 \\
&H2T~\cite{VU2023handcrafted}&  0.639 $\pm$ 0.11 & 0.662 $\pm$ 0.09 & 0.583 $\pm$ 0.03 & 0.603 $\pm$ 0.04 \\
&OT~\cite{mialon2021a} & 0.622 $\pm$ 0.09 & 0.687 $\pm$ 0.08 & 0.641 $\pm$ 0.02 & 0.495 $\pm$ 0.04 \\
&$\ours_{\text{WA}}$ + lin. &  0.647 $\pm$ 0.12 & 0.654 $\pm$ 0.07 & 0.461 $\pm$ 0.01 & 0.482 $\pm$ 0.06 \\
&$\ours_{\text{All}}$ + lin. & 0.645 $\pm$ 0.07 & 0.672 $\pm$ 0.06 & 0.568 $\pm$ 0.05 & 0.623 $\pm$ 0.07 \\
&$\ours_{\text{All}}$ + MLP &  0.665 $\pm$ 0.10 & 0.685 $\pm$ 0.06 & 0.653 $\pm$ 0.04 & 0.634 $\pm$ 0.04 \\
\bottomrule
\end{tabular}
\label{tab:ablation_loss}
\end{table*}

\clearpage

\end{document}